\documentclass{article}

 \usepackage[main, final, nonatbib]{neurips_2025}

\usepackage{enumitem}
\usepackage[utf8]{inputenc} 
\DeclareUnicodeCharacter{0302}{}
\usepackage[T1]{fontenc}    
\usepackage{hyperref}       
\usepackage{url}            
\usepackage{booktabs}       
\usepackage{amsfonts}       
\usepackage{nicefrac}       
\usepackage{microtype}      
\usepackage{xcolor}         
\usepackage{graphicx}
\usepackage{multirow}
\usepackage{subcaption}
\usepackage{multirow}
\usepackage{float}
\usepackage[table]{xcolor} 
\usepackage{gensymb}
\usepackage{amsmath}
\usepackage[sorting=none, maxbibnames=99]{biblatex}

\def\eg{\emph{e.g.}} 
\def\ie{\emph{i.e.}}

\def\etal{\emph{et al.}}

\title{WeatherPrompt: Multi-modality Representation Learning for All-Weather Drone Visual Geo-Localization}

\author{
  Jiahao Wen$^{1}$ \quad Hang Yu$^{1}$\thanks{Corresponding author.} \quad Zhedong Zheng$^{2}$ \\
  $^{1}$School of Computer Engineering and Science, Shanghai University, China \\
  $^{2}$Faculty of Science and Technology and Institute of Collaborative Innovation, University of Macau, China \\
  \texttt{\{wenjh,yuhang\}@shu.edu.cn, zhedongzheng@um.edu.mo}
}

\begin{document}

\maketitle

\begin{abstract}

Visual geo-localization for drones faces critical degradation under weather perturbations, \eg, rain and fog, where existing methods struggle with two inherent limitations: 1) Heavy reliance on limited weather categories that constrain generalization, and 2) Suboptimal disentanglement of entangled scene-weather features through pseudo weather categories.
We present WeatherPrompt, a multi-modality learning paradigm that establishes weather-invariant representations through fusing the image embedding with the text context. 
Our framework introduces two key contributions: First, a Training-free Weather Reasoning mechanism that employs off-the-shelf large multi-modality models to synthesize multi-weather textual descriptions through human-like reasoning. It improves the scalability to unseen or complex weather, and could reflect different weather strength. 
Second, to better disentangle the scene and weather features, we propose a multi-modality framework with the dynamic gating mechanism driven by the text embedding to adaptively reweight and fuse visual features across modalities. The framework is further optimized by the cross-modal objectives, including image-text contrastive learning and image-text matching, which maps the same scene with different weather conditions closer in the representation space. 
Extensive experiments validate that, under diverse weather conditions, our method achieves competitive recall rates compared to state-of-the-art drone geo-localization methods. Notably, it improves Recall@1 by 13.37\% under night conditions and by 18.69\% under fog and snow conditions. Our code is available at https://github.com/Jahawn-Wen/WeatherPrompt.
\end{abstract}


\section{Introduction}

Drone visual geo-localization aims to match drone-view image with corresponding satellite views, supporting critical applications such as disaster response, urban surveillance, search-and-rescue, and environmental monitoring~\cite{Zheng_Wei_Yang_202001,Wang_Zheng_Yan_Zhang_Sun_Zheng_Yang_202202, Lin_Zheng_Zhong_Luo_Li_Yang_Sebe_202203,968495006, zhu2022demo87, wang2024sparse88}.
However, variable weather conditions such as rain, fog and snow introduce noise, occlusions and low visibility, which severely distort image features~\cite{feng2024multi04,sindagi2020prior05,liu2022image07} and leading conventional localization methods to suffer drastic performance degradation under extreme weather. 
Recent advances in cross-modal retrieval show that integrating natural language descriptions can substantially enhance the discrimination power of vision models, allowing better generalization in complex or ambiguous scenarios~\cite{yang2023towards08,yan2024dual09,yan2023clip10,vendrow2024inquire11}. Despite this progress, leveraging textual guidance for cross-weather drone geo-localization remains largely underexplored, especially considering the nuanced and dynamic nature of weather conditions encountered in the field. The ability of text to capture complex semantics and fine-grained details~\cite{li2022fine64, chow2024unified65} offers a promising avenue for cross-weather generalization. Addressing the all-weather visual geo-localization task presents two primary challenges:
(1) Limited Weather Labeling. Existing approaches typically perform domain-specific fine-tuning on a limited set of predefined weather labels (\eg, sunny, rainy). This closed-set paradigm fails to capture the continuous and combinatorial nature of real-world weather, thereby limiting model generalization to unseen or mixed conditions and preventing the exploitation of richer weather semantics.
(2) Scene–Weather Feature Entanglement. Existing methods directly inject coarse pseudo-weather labels (\eg, rain, fog) into visual representations during training, leading to a severe entanglement between scene semantics and weather disturbances. Consequently, the model learns suboptimal representations under mixed or unseen weather conditions, fails to disentangle scene content from weather noise, and is severely limited in cross-weather generalization.

\begin{figure}[t]
    \centering
    \includegraphics[width=0.9\linewidth]{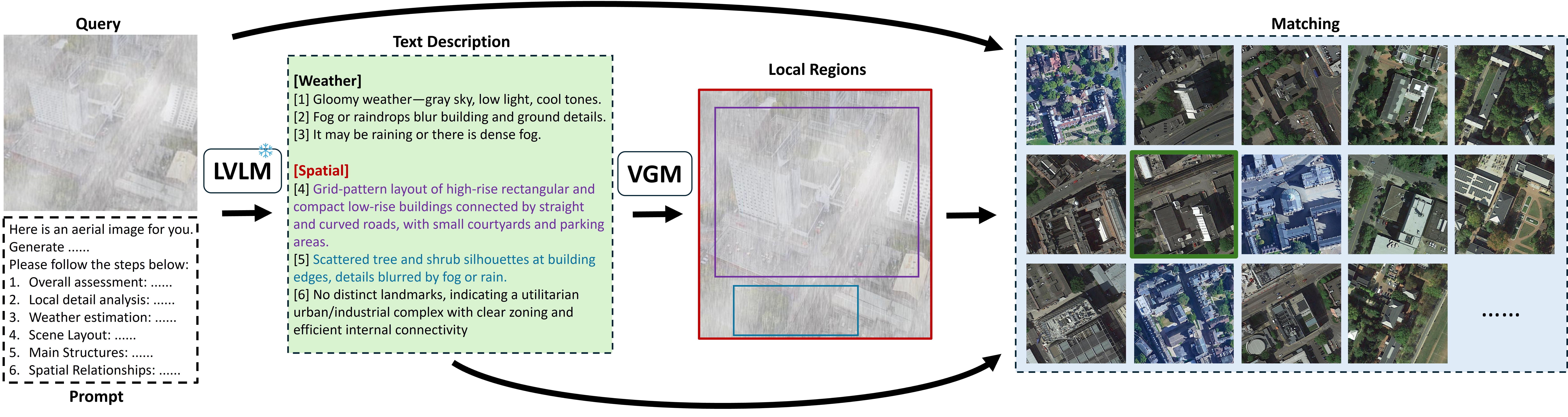}
    \caption{\textbf{Example of the proposed Chain-of-Thought description and matching.}
Our framework generates structured weather and spatial Text Description via stepwise reasoning.
We leverage Off-the-shelf Visual Grounding Model (VGM), \ie, XVLM~\cite{xvlm36} to extract local region cues, which are integrated to further refine the matching process.
Finally, we match images using weather description, global scene layout, and local region semantics to retrieve the corresponding satellite-view image. 
}
    \label{fig1}
    \vspace{-.20in}
\end{figure}

For the first limitation, we propose a training-free weather reasoning mechanism leveraging off-the-shelf large multimodal models~\cite{Qwen2.5-VL12} (see Fig.~\ref{fig1}). Specifically, we employ chain-of-thought (CoT)~\cite{wei2022chain13} prompting to automatically generate rich, step-by-step natural language weather descriptions for each geographical scene using a single randomly sampled drone-view image.  This strategy circumvents manual description and expert controlled, enabling large-scale collection of high-quality and diverse multimodal samples at significantly reduced costs.  Moreover, the stepwise reasoning introduced by the CoT prompts ensures semantic accuracy and formatting consistency across generated captions, further enhancing the reliability and usability of descriptions.  This ultimately leads to improved generalization for complex and unseen weather conditions.
To address the second challenge, we propose a multimodal framework equipped with a text embedding-driven dynamic gating mechanism to adaptively reweight and fuse visual features, effectively disentangling scene and weather attributes. Specifically, during training, the framework jointly optimizes multimodal objectives including image-text contrastive (ITC) loss and image-text matching (ITM) loss, aligning drone images with generated weather-aware captions.   Additionally, a localized alignment loss is introduced to explicitly enforce consistency between annotated visual regions and textual descriptions across multiple granularities, encouraging the visual encoder to learn robust and weather-invariant scene representations. At inference, visual and textual embeddings are extracted in parallel, and the textual embeddings dynamically modulate visual features via the gating mechanism. The resulting multimodal representation is directly fed into a classification head for localization prediction, avoiding any additional online fine-tuning or dedicated parameter sets for different weather conditions. This design significantly reduces structural complexity and improves the deployment efficiency on resource-constrained platforms.
The main contributions are as follows:
\begin{itemize}[label={},leftmargin=*,align=left]
     \item[$\bullet$] \textbf{Training-Free Weather Reasoning}: We pioneer automatic weather semantics extraction through Large Vision Language Models (LVLMs) with chain-of-thought prompting, eliminating manual descriptions. Our hierarchical reasoning mechanism integrates continuous weather priors with spatial-object analysis, enabling weather-adaptive caption generation and scalable dataset construction.
     \item[$\bullet$] \textbf{Semantic Disentanglement via Language Guidance}: We devise a text-driven framework achieving scene-weather disentanglement through: (1) Multi-granularity alignment of visual features with continuous textual weather embeddings, (2) Region-level semantic consistency enforcement, (3) Dynamic textual gating for weather-invariant representations.
    \item[$\bullet$] \textbf{State-of-the-Art Generalization}: The proposed method has achieved average 87.72\% Recall@1 on University-1652 and 83.1\% on SUS-200 over 10 different weather conditions. For unseen weather combinations (\ie, Dark+Rain+Fog), ours still arrives at 72.15\% AP, validating unprecedented cross-domain generalization.
\end{itemize}


\section{Related Work}
\label{Related Work}

\noindent\textbf{Cross-view Geolocalization.} Cross-view geo-localization aims to match images captured from different viewpoints with their corresponding geographic locations~\cite{Zheng_Wei_Yang_202001, berton2022deep14, jin2017learned15, chu2024towards16}. Early approaches relied on hand-crafted local features such as SIFT~\cite{cruz2012scale17} and SURF~\cite{bay2006surf18}, as well as global descriptors like VLAD~\cite{zhang2021vector19} and Fisher Vector~\cite{sanchez2013image20}, often combined with RANSAC~\cite{barath2022learning21} for geometric verification; however, they remain brittle under large viewpoint and illumination changes~\cite{Castaldo_Zamir_Angst_Palmieri_Savarese_201522,Lin_Belongie_Hays_201323}.
With the advent of deep learning, deep metric learning frameworks based on global or part-based contrastive objectives have become dominant~\cite{wang2024Muse24,Workman_Souvenir_Jacobs_201525,Lin_YinCui_Belongie_Hays_201526, vepa2024integrating66, choi2023depth68, zheng2017discriminatively}. These approaches employ triplet or InfoNCE losses to train end-to-end embeddings and integrate global pooling with spatial attention or multi-region partitioning strategies~\cite{yang2021cross27,rodrigues2022global28}. Representative examples include the multi-part partitioning scheme~\etal~\cite{Wang_Zheng_Yan_Zhang_Sun_Zheng_Yang_202202}, the keypoint attention module~\cite{Lin_Zheng_Zhong_Luo_Li_Yang_Sebe_202203}, the content-aligned Transformer architecture~\cite{dai2021transformer29}, self-attention positional encoding by Yang~\etal~\cite{yang2021cross27}, the dual-path fusion network~\cite{rodrigues2022global28}, and Bird's Eye View (BEV)~\cite{ju2024video2bev}, all of which substantially enhance cross-view feature alignment.
Recent research has begun to address the impact of image degradations such as low-light, motion blur, and synthetic fog, often by using data augmentation, domain adaptation, or cross-modal transformers~\cite{tan2023mapd31, ma2022both30, zhang2023cross32, zhu2022transgeo33, qu2024lush69, chen2024restoreagent70}. However, most existing methods still depend on a limited set of discrete weather labels, restricting generalization to unseen or complex conditions. In contrast, our approach introduces a training-free, all-weather text-guided representation learning framework that leverages open-set weather descriptions to overcome these limitations.

\noindent\textbf{Multi-modality Alignment.} 
In this work, we address weather-aware text-guided representation learning, where the goal is to retrieve drone-view images based on fine-grained weather-related textual cues. Recent advances in vision–language alignment, such as CLIP~\cite{radford2021learning34}, BLIP~\cite{li2022blip35}, and XVLM~\cite{xvlm36}, have established powerful contrastive pre-training and cross-modal attention mechanisms, but previous studies mainly target static semantics or rigid spatial relations~\cite{zhu2024mvp37, chen2024rsprompter38, yuan2023parameter39}. Additional efforts on adaptive fusion and region-word alignment~\cite{zheng2020dual40, wang2019camp41, chen2020uniter43, li2020oscar44} have improved retrieval, but remain limited by closed vocabularies and overlook dynamic, fine-grained weather semantics.
To address these gaps, we propose a framework that generates open-set weather descriptions via Chain-of-Thought prompting and applies text-driven dynamic gating for adaptive feature modulation, achieving robust cross-modal alignment under diverse weather conditions.

\noindent\textbf{Large Vision Language Models for Vision via Prompting.}
Large Vision Language Models (LVLMs) such as GPT‑3/4~\cite{brown2020language45,achiam2023gpt46} and Qwen~\cite{Qwen2.5-VL12} have recently been applied to vision tasks using prompt engineering. Approaches like VisualGPT~\cite{chen2022visualgpt47} and MM‑CoT~\cite{zhang2023multimodal48} leverage Chain-of-Thought (CoT) prompts~\cite{wei2022chain13} to elicit stepwise reasoning for visual question answering and captioning. However, scaling LVLMs to cross-view, multi-weather geo-localization remains challenging: free-form text often suffers from hallucinations and lacks semantic or structural consistency~\cite{sriramanan2024llm49, jiang2024hallucination50, kim2024exploiting51}. Prior works also overlook prompt design tailored for robust, multi-weather, multi-scale cross-modal alignment.
To address this, we introduce the first CoT-driven description pipeline for multi-weather drone-to-satellite geo-localization. Our structured prompts regularize LVLMs' outputs, enabling scalable generation of high-quality, open-set weather descriptions to advance large-scale vision–language alignment.

\begin{figure}[t]
    \centering
    \includegraphics[width=\linewidth]{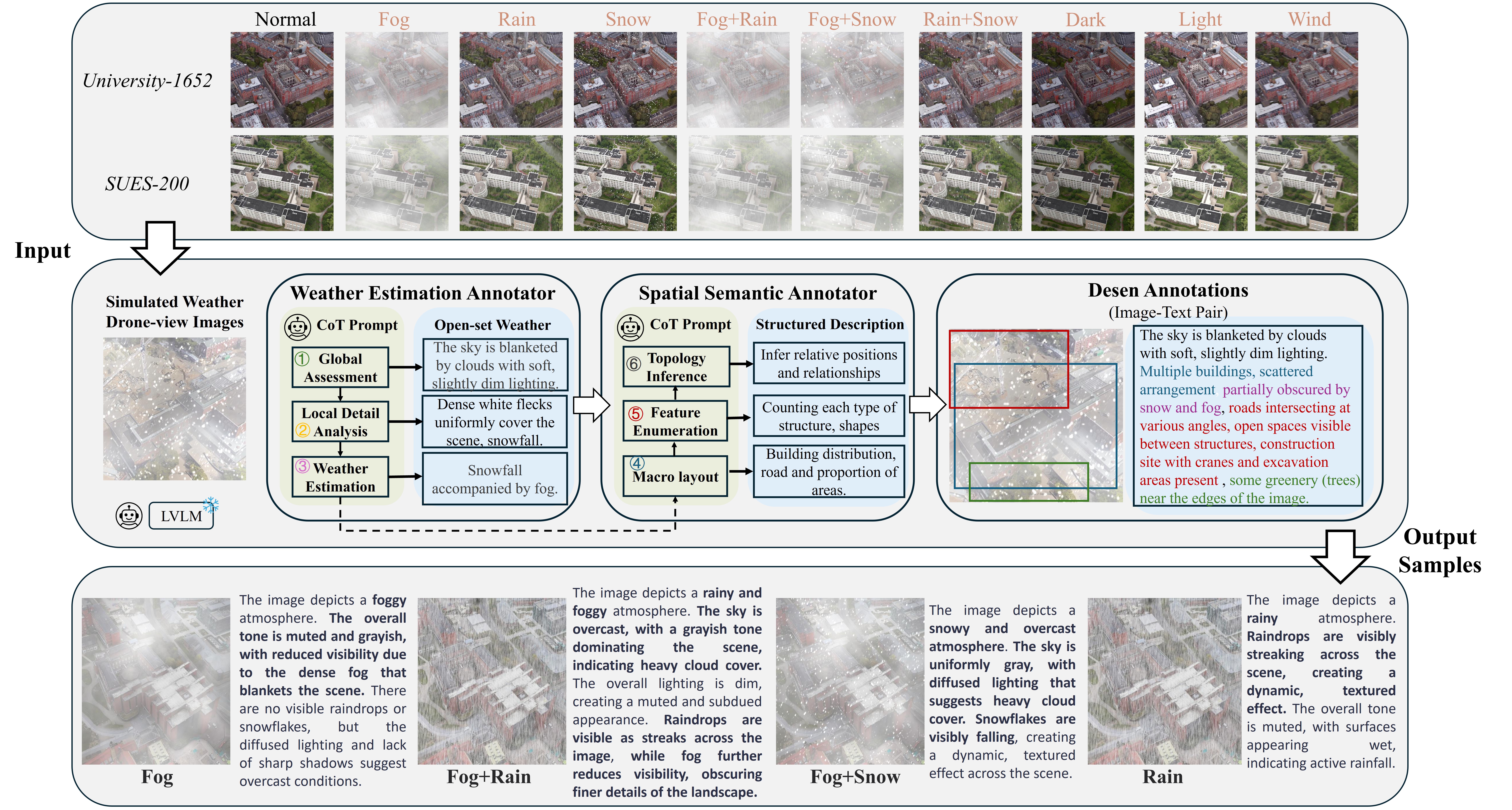}
    \vspace{-.10in}
    \caption{\textbf{The proposed training-free weather reasoning mechanism}. We synthesize drone-view images with diverse weather conditions based on the University-1652 and SUES-200 datasets, covering complex scenarios such as fog, rain, snow, and nighttime. For each synthesized image, we first employ stepwise Chain-of-Thought prompting to generate open-set weather descriptions, including global assessment, local detail analysis, and weather inference. Guided by the inferred weather prior, we then sequentially reason about the scene’s macro layout, structural elements, and topological relationships, ultimately producing high-quality, structured image–text pairs.}
    \label{fig2}
    \vspace{-.20in}
\end{figure}





\section{Method}
\label{Method}

\subsection{Open-Weather Description}
As shown in Fig.~\ref{fig2}, we present an overview of our multi-weather drone-view image captioning pipeline. To minimize description redundancy and mitigate overfitting, our approach begins by randomly sampling a single representative drone-view image from each geographical region, forming concise yet diverse image–text pairs.
Notably, Large Vision Language Models (LVLMs) typically generate weather labels or captions intuitively, which may omit critical visual cues and exhibit semantic inconsistency. To address this limitation, we explicitly divide the captioning process into two sequential phases: a weather estimation phase and a spatial semantics phase, guided by carefully designed Chain-of-Thought (CoT) prompts. Inspired by human hierarchical visual reasoning, our CoT prompting strategy enforces a rigorous three-step reasoning procedure: global perception, local analysis, and comprehensive synthesis.

In the weather estimation phase, the LVLMs first assesses global visibility to quantify observable range, subsequently identifies localized atmospheric indicators such as rain-streak reflections or fog diffusion patterns, and finally integrates these global and local cues to assign an accurate weather category.   
This structured inference process significantly alleviates the ambiguity inherent to single-shot captioning methods, establishing a reliable weather-conditioned prior for the following stage.

In the subsequent spatial semantics phase, conditioned explicitly on the inferred weather semantic, the model rapidly identifies macro-level layout features, including the spatial distribution of buildings, orientations of roads, and proportion of open spaces.   Then, it captures micro-level structural details such as object counts, shapes, and local spatial arrangements. 
Ultimately, the pipeline synthesizes these cues by reasoning about relative positions and topological relationships to generate a structured textual description.   This final output aligns comprehensively with weather semantics, spatial structure, and detailed scene attributes, ensuring robust, consistent, and high-quality multimodal descriptions.


\textbf{Single-image Sampling.}
Since existing benchmarks (\eg, University-1652) are partitioned by geographic regions, we randomly select only one drone-view image per region as a representative to avoid redundancy. Given the prohibitive cost of obtaining real drone-view imagery across diverse weather conditions, we utilize the imgaug~\cite{imgaug67} library to synthesize realistic weather variations. By parametrically adjusting visibility and occlusion effects such as rain, fog, and snow, we generate high-fidelity, diverse meteorological scenarios that closely resemble real-world conditions.

\textbf{Weather Estimation Phase.}
Given a drone-view image with synthetically generated weather, we apply a pretrained large multimodal model~\cite{Qwen2.5-VL12} for automatic weather description. To mitigate hallucinations, vague phrasing, and inconsistent terminology typical of single-shot captioning, we adopt a Chain-of-Thought prompting strategy inspired by human visual perception. Specifically, we first quantify global visibility, then identify local meteorological cues such as rain-streak reflections or fog diffusion, and finally integrate global and local evidence to determine a reliable weather label. Any description that lacks explicit visibility information, meteorological cues, or contains uncertain terminology (\eg, “possibly” or “uncertain”) is automatically rejected and regenerated. This three-step reasoning pipeline ensures accurate, consistent, and reliable weather descriptions, thus providing a robust prior for the subsequent spatial semantics phase.

\textbf{Spatial Semantics Phase.}
Conditioned on the inferred weather prior, we further generate detailed, scene-level textual descriptions essential for precise vision–language alignment. Initially, the model conducts a global scan to capture macro-level structures, including building distributions, road orientations, and proportions of open areas. Subsequently, it enumerates fine-grained elements, accurately counting structural entities, describing shapes, and detailing local spatial arrangements. Lastly, it synthesizes macro-layout and micro-level information to explicitly infer relative positions and topological relationships, producing structured, natural-language scene descriptions. These descriptions serve as fine-grained semantic supervision for downstream multimodal alignment.


\textbf{Discussion.}
Existing cross-view geo-localization methods primarily rely on image-based retrieval. While some recent studies~\cite{chu2024towards16, ye2024cross52} integrate textual descriptions, these approaches remain limited to coarse, scene-level labels and largely neglect weather semantics, resulting in compromised robustness under diverse weather conditions. In contrast, we introduce a training-free, CoT-driven pipeline for automatically generating multi-weather semantic descriptions. By employing large multimodal models guided by Chain-of-Thought prompting, we move beyond predefined discrete weather categories and achieve robust generalization to unseen and complex meteorological conditions. Furthermore, our framework systematically annotates both macro-level spatial layouts and micro-level structural attributes, embedding essential spatial cues into the image–text pairs. Relying solely on pretrained multimodal models without manual description, our approach efficiently produces large-scale, semantically consistent multimodal data, significantly facilitating downstream cross-weather multimodal representation learning.

\begin{figure}[t]
    \centering
    \includegraphics[width=\linewidth]{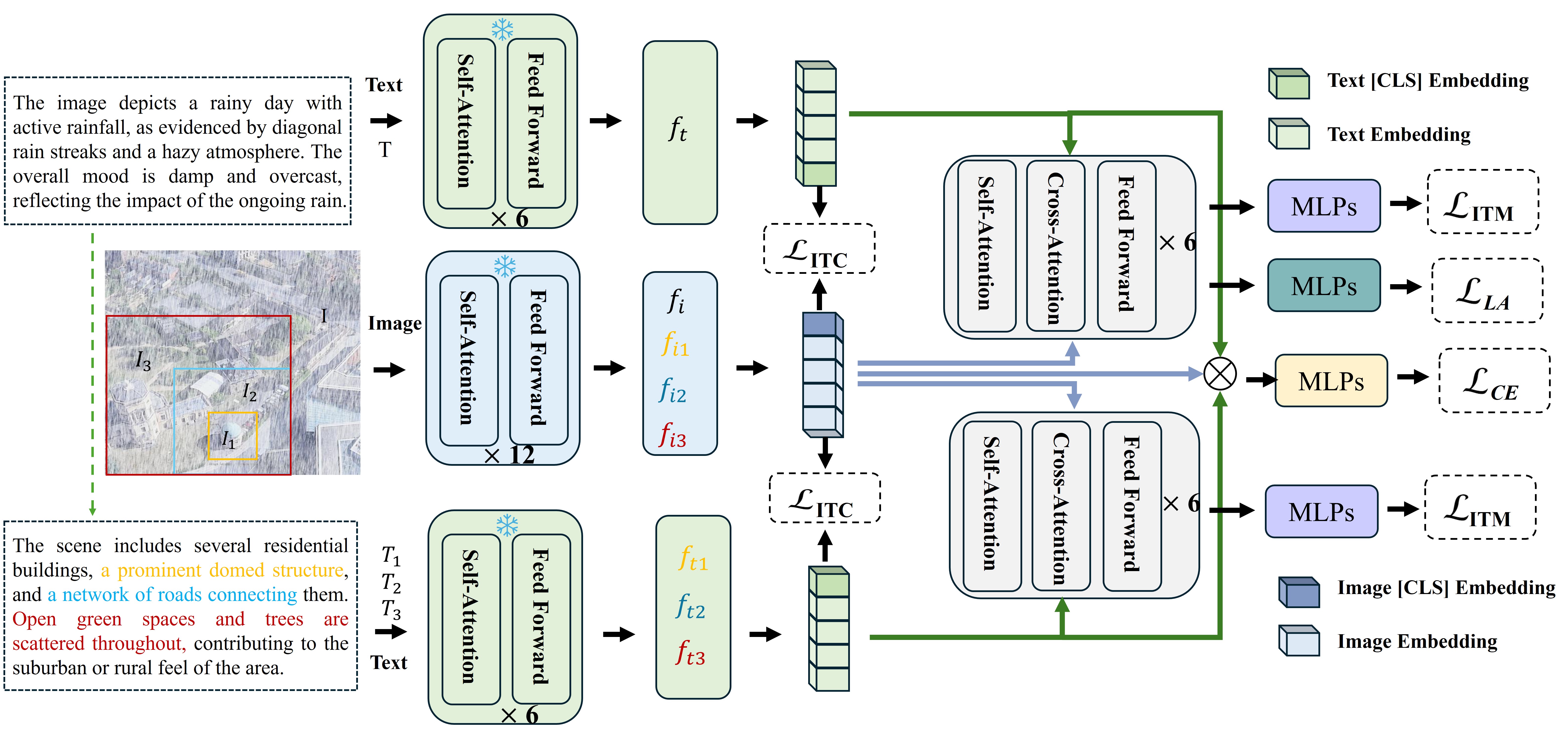}
    \vspace{-.15in}
    \caption{\textbf{The proposed multimodal alignment framework.} Our model extracts global and local features from drone images and multi-step weather captions, performs multi-granular image-text alignment, and dynamically fuses modalities via weather-driven gating for robust geo-localization.}
    \label{fig3}
    \vspace{-.10in}
\end{figure}

\subsection{Multimodal Alignment Model}
We introduce a training-free, all-weather text-guided representation learning framework (see Fig.~\ref{fig3}) that aligns weather semantics injected via text prompts with visual features to produce consistent, discriminative cross-view embeddings across diverse meteorological conditions. The framework consists of a pretrained visual encoder, a text encoder, and a cross-modal fusion module, capped by a lightweight classification head. During inference, the text embedding parametrizes a gating mechanism that adaptively modulates the visual feature channels; the fused multimodal representation is then passed through a single-layer MLP to predict the geographic location.

\textbf{Weather-Driven Channel Gating.} Let \(f_{\mathrm{I}}\in\mathbb{R}^{B\times D}\) and \(f_{\mathrm{T}}\in\mathbb{R}^{B\times D}\) represent the normalized visual and textual embeddings for a batch of $B$ samples. We utilize text embeddings to generate an adaptive channel-wise gating vector \(g\in(0,1)^{B\times D}\):
\begin{equation}\label{eq1}
\begin{aligned}
z &= \mathrm{ReLU}\bigl(f_{\mathrm{T}}W_1^{(gate)\top} + b_1^{(gate)}\bigr)\in\mathbb{R}^{B\times D/r},\quad
g &= \sigma\bigl(zW_2^{(gate)\top} + b_2^{(gate)}\bigr)\in(0,1)^{B\times D},
\end{aligned}
\end{equation}
where \(W_1^{(gate)}\), \(W_2^{(gate)}\), \(b_1^{(gate)}\), \(b_2^{(gate)}\) are learnable parameters, and \(r\) is the reduction ratio. 
The fused multimodal feature is computed by
\begin{equation}
\label{eq2}
f_{\mathrm{fuse}} = g \odot f_{\mathrm{I}} + (1 - g)\odot f_{\mathrm{T}}
\quad\in\mathbb{R}^{B\times D},
\end{equation}
followed by a classification head for geo-localization.
By dynamically modulating visual channels conditioned on textual weather semantics, our approach yields discriminative and weather-robust embeddings with minimal computational overhead.

\textbf{Image-Text Contrastive.}
Given paired samples \(\{(I_i, T_i)\}_{i=1}^B\), we compute the similarity matrix \(
S_{ij} = \frac{I_i^\top T_j}{\tau}\;\in\mathbb{R}^{B\times B}\) between normalized visual embeddings \(I_i\) and textual embeddings \(T_j\), where \(\tau\) is a learnable temperature parameter. We then convert the diagonal entries into retrieval probabilities by applying softmax over rows and columns:
\begin{equation}\label{eq4}
\begin{aligned}
p_{I\to T}^{(i)} &= \frac{\exp(S_{ii})}{\sum_{j=1}^B \exp(S_{ij})}, 
\quad
p_{T\to I}^{(i)} &= \frac{\exp(S_{ii})}{\sum_{j=1}^B \exp(S_{ji})}.
\end{aligned}
\end{equation}

The contrastive loss is defined as:
\begin{equation}
\label{eq5}
\mathcal{L}_{\mathrm{ITC}}
= -\frac{1}{2B}\sum_{i=1}^B \Bigl[\log p_{I\to T}^{(i)} + \log p_{T\to I}^{(i)}\Bigr],
\end{equation}
enforcing global semantic alignment between the visual and textual modalities.

\textbf{Image-Text Matching.}
To further enhance fine-grained discriminability, we introduce an image-text matching loss utilizing hard negatives. Given the similarity matrix \(S\), for each image \(I_i\) and its paired textual \(T_i\)), we select the highest non–diagonal similarity entry as its hard negative:  
\(
T_i^- = \arg\max_{j\neq i} S_{ij},
I_i^- = \arg\max_{j\neq i} S_{ji}.
\)
This forms \(2B\) hard negative pairs \((I_i, T_i^-)\) and \((I_i^-, T_i)\), and \(B\) positive pairs \((I_i, T_i)\), resulting in $3B$ pairs in total. Each pair is passed through the cross-modal encoder, and we extract the CLS embedding \(h_k\) which is fed to a binary classifier \(f_{\text{match}}\). The matching probability is \(p_k = \sigma\bigl(f_{\text{match}}(h_k)\bigr),\)
where \(\sigma\) is the sigmoid function. Denoting the ground-truth label \(y_k = 1\) for positive pairs and \(y_k = 0\) for negatives, the matching loss is defined as
\begin{equation}
\label{eq:itm}
\mathcal{L}_{\mathrm{ITM}}
= -\frac{1}{3B}\sum_{k=1}^{3B}\Bigl[y_k\log p_k + (1-y_k)\log(1-p_k)\Bigr],
\end{equation}
which guides the model to hear subtle semantic distinctions between closely related pairs.

\textbf{Localized Alignment Module.}
To enable text-driven fine-grained visual localization, the model responds to both global descriptions $T$ and the region hints $T_1, T_2, T_3$. For the j-th concept , we extract its [CLS] embedding \(x_{\mathrm{cls}}^j \in \mathbb{R}^d.\) from the cross-encoder, which is projected to normalized coordinate vector via a dedicated two‐layer MLP:
\begin{equation}
\hat{l}^j = \sigma\bigl( W_2^{(loc)}\,\mathrm{GELU}(W_1^{(loc)}\,x_{\mathrm{cls}}^j + b_1^{(loc)}) + b_2^{(loc)} \bigr) \in [0,1]^4
\end{equation}

where \(W_1^{(loc)}\in\mathbb{R}^{2d\times d}\), \(W_2^{(loc)}\in\mathbb{R}^{4\times 2d}\), \(b_1^{(loc)}\in\mathbb{R}^{2d}\), \(b_2^{(loc)}\in\mathbb{R}^4\). The resulting vector \(\hat{l}^j=(\hat c_x,\hat c_y,\hat w,\hat h)\) encodes the region center and size in normalized coordinates.
The localized alignment loss supervises both overlap and regression accuracy:
\begin{equation}
\mathcal{L}_{\mathrm{LA}}
= \mathbb{E}_{(I,T^j)\sim\mathcal{D}}
\Bigl[
\underbrace{1 - \mathrm{IoU}\bigl(l^j,\hat{l}^j\bigr)}_{\text{Overlap Loss}}
\;+\;
\underbrace{\lVert l^j - \hat{l}^j\rVert_{1}}_{\text{L1 Loss}}
\Bigr],
\label{eq:local-align-loss}
\end{equation}
where \(l^j=(c_x,c_y,w,h)\) is the ground‐truth region, \(\hat{l}^j\) its prediction, and \(\mathrm{IoU}(\cdot,\cdot)\) computes the Intersection‐over‐Union.

\textbf{Classification Module.}
After multimodal fusion, we employ a single-layer MLP classification head to predict the geographic location. Let the fused feature for sample \(b\) be \(z_b \in \mathbb{R}^D\), and classifier output logits 
\(
o_b = W_{\mathrm{clf}} z_b + b_{\mathrm{clf}} \in \mathbb{R}^C,
\)
where \(W_{\mathrm{clf}}\in\mathbb{R}^{C\times D}\) and \(b_{\mathrm{clf}}\in\mathbb{R}^C\). Given the ground-truth location label \(y_b\in\{1,\dots,C\}\), the softmax cross-entropy loss is:
\begin{equation}
\label{eq6}
\mathcal{L}_{\mathrm{CE}}
= -\frac{1}{B}\sum_{b=1}^B 
\log\frac{\exp\bigl(o_{b,y_b}\bigr)}
{\sum_{c=1}^C \exp\bigl(o_{b,c}\bigr)}.
\end{equation}

\textbf{Optimization Objectives.} Our total loss \(L_{total}\) is defined as:
\begin{equation}
\label{eq7}
\mathcal{L}_{total} = \mathcal{L}_{ITC} + \mathcal{L}_{ITM} + \mathcal{L}_{LA} + \mathcal{L}_{CE}.
\end{equation}

This unified objective enables the model to learn fine-grained, weather-invariant representations, substantially improving cross-weather generalization.

\section{Experiment}
\label{experiment}
\textbf{Implementation Details.} We adopt XVLM~\cite{xvlm36} as the backbone, which is pre-trained on 4M image–caption pairs, integrates BERT~\cite{devlin2019bert54} as the text encoder and Swin Transformer~\cite{liu2021swin55} as the image encoder. The model is optimized using stochastic gradient descent (SGD)~\cite{robbins1951stochastic53} with momentum $0.9$ and weight decay $0.0005$. Training 210 epochs, with the learning rate reduced by 0.1 at epoch 120 and by 0.01 at epoch 180. We resize input images to $384 \times 384$ pixels and divide them into $32 \times 32$ patches. During training, satellite‐view images are augmented via random cropping and horizontal flipping, for drone‐view images, we first apply style transformations using the imgaug~\cite{imgaug67} library and then perform the same random crop and flip augmentations. At test time, we compute the Euclidean distance between query and candidate embeddings to measure similarity. All experiments have been implemented in PyTorch~\cite{paszke2019pytorch56} and conducted on a single NVIDIA RTX A6000 GPU, with an average inference time of $0.024$s per query.

\textbf{Dateset.}
\textbf{University-1652}~\cite{Zheng_Wei_Yang_202001} is a large-scale cross-view geo-localization dataset comprising images from 1,652 university locations. Each location is represented by satellite, drone, and ground-level images, with 54 drone-view and 1 satellite-view images per building, as well as street-view imagery. The dataset is split into 701 training and 951 test buildings, with no overlap between train and test sets.
\textbf{SUES-200}~\cite{zhu2023sues57} contains multi-view drone and satellite images from 200 locations in Shanghai, encompassing diverse urban scenes, parks, lakes, and public buildings. Drone images are captured from multiple altitudes (150–300m) to simulate varied real-world conditions.


\begin{table*}[t]
\centering
\resizebox{\textwidth}{!}{%
\begin{tabular}{l|cc|cc|cc|cc|cc|cc|cc|cc|cc|cc|cc}
\hline
\multirow{2}{*}{Method} & \multicolumn{2}{c|}{Normal} & \multicolumn{2}{c|}{Fog} & \multicolumn{2}{c|}{Rain} & \multicolumn{2}{c|}{Snow} & \multicolumn{2}{c|}{Fog+Rain} & \multicolumn{2}{c|}{Fog+Snow} & \multicolumn{2}{c|}{Rain+Snow} & \multicolumn{2}{c|}{Dark} & \multicolumn{2}{c|}{Over-exp} & \multicolumn{2}{c|}{Wind} & \multicolumn{2}{c}{Mean} \\
 & R@1 & AP & R@1 & AP & R@1 & AP & R@1 & AP & R@1 & AP & R@1 & AP & R@1 & AP & R@1 & AP & R@1 & AP & R@1 & AP & R@1 & AP \\
\hline
\multicolumn{12}{r}{\textbf{Drone $\rightarrow$ Satellite}} \\
\hline
Zheng~\etal \cite{Zheng_Wei_Yang_202001} \textcolor{gray}{[backbone]} & 67.83 & 71.74 & 60.97 & 65.23 & 60.29 & 64.61 & 55.58 & 60.09 & 54.75 & 59.40 & 44.85 & 49.78 & 57.61 & 62.03 & 39.70 & 44.65 & 51.85 & 56.75 & 58.28 & 62.83 & 55.17 & 59.71 \\
ResNet-101 \cite{He_Zhang_Ren_Sun_201658} \textcolor{gray}{[backbone]} & 70.07 & 73.04 & 63.87 & 68.22 & 63.34 & 67.59 & 59.75 & 64.15 & 57.45 & 62.12 & 48.31 & 53.28 & 60.25 & 64.68 & 46.12 & 51.02 & 56.34 & 61.23 & 62.13 & 66.63 & 58.76 & 63.29 \\
DenseNet121 \cite{huang2017densely59} \textcolor{gray}{[backbone]} & 69.48 & 73.26 & 64.25 & 68.47 & 63.47 & 67.64 & 59.29 & 63.70 & 59.68 & 64.13 & 50.41 & 55.20 & 60.21 & 64.57 & 48.57 & 53.41 & 54.04 & 58.88 & 60.74 & 65.14 & 59.01 & 63.44 \\
Swin-T \cite{liu2021swin55} \textcolor{gray}{[backbone]} & 69.27 & 73.18 & 66.46 & 70.52 & 65.44 & 69.60 & 61.79 & 66.23 & 63.96 & 68.21 & 56.44 & 61.07 & 62.68 & 67.02 & 50.27 & 55.18 & 55.46 & 60.29 & 63.81 & 68.17 & 61.56 & 65.95 \\
IBN-Net \cite{Pan_Luo_Shi_Tang_201860} \textcolor{gray}{[backbone]} & 72.35 & 75.85 & 66.68 & 70.64 & 67.95 & 71.73 & 62.77 & 66.85 & 62.64 & 66.84 & 51.09 & 55.79 & 64.07& 68.13 & 50.72 & 55.53 & 57.97 & 62.52 & 66.73 & 70.68 & 62.30 & 66.46 \\
LPN \cite{Wang_Zheng_Yan_Zhang_Sun_Zheng_Yang_202202} \textcolor{blue}{[TCSVT’21]} & 74.33 & 77.60 & 69.31 & 72.95 & 67.96 & 71.72 & 64.90 & 68.85 & 64.51 & 68.52 & 54.16 & 58.73 & 65.38 & 69.29 & 53.68 & 58.10 & 60.90 & 65.27 & 66.46 & 70.35 & 64.16 & 68.14 \\
Sample4Geo\textsuperscript{*} \cite{deuser2023sample4geo61} \textcolor{blue}{[ICCV’23]} & \textbf{92.70} & \textbf{93.85} & \textbf{88.70} & \textbf{90.55} & 62.44 & 66.17 & 52.76& 57.24 & 52.70& 56.77& 19.79& 23.16& 38.19& 42.33& 46.34& 49.91& 75.77& 78.90 & \textbf{81.54} & \textbf{87.34} & 61.10 & 64.32 \\  
Safe-Net\textsuperscript{*} \cite{lin2024self62} \textcolor{blue}{[TIP’24]} & 86.98 & 88.85 & 82.12 &  86.10 &  67.13 & 68.90  & 60.50  & 63.01  & 54.80  & 58.73  & 32.12  &  39.77 & 25.83  & 26.40  & 41.10  & 44.13  & 69.87  &  71.15 & 74.32  & 76.58  & 60.48  & 63.36  \\
CCR\textsuperscript{*} \cite{du2024ccr63} \textcolor{blue}{[TCSVT’24]} & 92.54 & 93.78 &  85.57 & 87.13 & 67.46  &  68.82 & 55.16  &  59.14 & 63.11  & 60.97  & 27.74  & 31.48  &  23.06 & 46.85  &  51.10 & 54.19  & \textbf{75.90}  & \textbf{79.16}  &  81.31 & 87.22  & 62.30  & 66.87  \\
MuSe-Net\textsuperscript{*} \cite{wang2024Muse24}\textcolor{blue}{[PR’24]} & 74.48 & 77.83 & 69.47 & 73.24 & 70.55 & 74.14 & 65.72 & 69.70 & 65.59 & 69.64 & 54.69 & 59.24 & 66.64 & 70.55 & 53.85 & 58.49 & 61.05 & 65.51 & 69.45 & 73.22 & 65.15 & 69.16 \\
\hline
\rowcolor{gray!15} 
Ours & 82.78 & 85.18 & 81.46 & 84.03 & \textbf{80.34} & \textbf{83.11} & \textbf{77.60} & \textbf{80.67} & \textbf{78.75} & \textbf{81.69} & \textbf{73.38} & \textbf{76.94} & \textbf{78.41} & \textbf{81.40} & \textbf{67.22} & \textbf{71.06} & 74.20 & 77.63 & 77.26 & 80.27 & \textbf{77.14} \textcolor{red}{($+11.99$)}& \textbf{80.20} \textcolor{red}{($+11.04$)}\\
\hline
\multicolumn{12}{r}{\textbf{Satellite $\rightarrow$ Drone}} \\
\hline
Zheng~\etal \cite{Zheng_Wei_Yang_202001} \textcolor{gray}{[backbone]} & 83.45 & 67.94 & 79.60 & 61.12 & 77.60 & 59.73 & 73.18 & 55.07 & 75.89 & 54.45 & 70.76 & 43.26 & 74.75& 56.44 & 69.47 & 39.25 & 72.18 &51.91 & 76.46 & 57.59 & 75.33 & 54.68 \\
ResNet-101 \cite{He_Zhang_Ren_Sun_201658} \textcolor{gray}{[backbone]} & 85.73 & 71.79 & 82.45 & 66.46 & 81.46 & 65.68 & 79.74 & 61.72 & 79.74 & 60.59 & 74.75 & 50.31 & 80.17 & 62.61 & 75.32 & 45.37 & 79.60 & 58.21 & 82.31 & 64.67 & 80.13 & 60.74 \\
DenseNet121 \cite{huang2017densely59} \textcolor{gray}{[backbone]} & 83.74 & 70.34 & 82.31 & 66.32 & 81.17 & 65.23 & 78.60 & 60.33 & 79.46 & 61.66 & 74.61 & 51.14 & 78.46 & 61.68 & 74.47 & 47.88 & 74.32 & 55.26 & 78.32 & 61.63 & 78.55 & 60.15 \\
Swin-T \cite{liu2021swin55} \textcolor{gray}{[backbone]} & 80.74 & 68.94 & 81.03 & 67.46 & 81.17 & 66.39 & 78.46 & 61.33 & 79.17 & 64.65 & 74.89 & 56.57& 78.89 & 63.49 & 75.61 & 48.43 & 76.60 & 56.57 & 78.74 & 64.45 & 78.53 & 61.83 \\
IBN-Net \cite{Pan_Luo_Shi_Tang_201860} \textcolor{gray}{[backbone]} & 86.31 & 73.54 & 84.59 & 67.61 & 84.74 & 69.03 & 80.88 & 64.44 & 83.31 & 63.71 & 77.89 & 52.14 & 83.02 & 65.74 & 78.46 & 50.77 & 79.46 & 58.64 & 84.02 & 67.94 & 82.27 & 63.36 \\
LPN \cite{Wang_Zheng_Yan_Zhang_Sun_Zheng_Yang_202202} \textcolor{blue}{[TCSVT’21]} & 87.02 & 75.19 & 86.16 & 71.34 & 83.88 & 69.49 & 82.88 & 65.39 & 84.59 & 66.28 & 79.60 & 55.19 & 84.17 & 66.26 & 82.88 & 52.05 & 81.03 & 62.24 & 84.14 & 67.35 & 83.64 & 65.08 \\
Sample4Geo\textsuperscript{*} \cite{deuser2023sample4geo61} \textcolor{blue}{[ICCV’23]} & \textbf{95.29} & 91.42 & \textbf{93.87} & \textbf{87.46} & 73.04 & 50.27 & 76.18 & 47.58 & 71.18 & 44.53& 52.21 & 16.21 & 64.48 & 32.38 & 77.03 & 45.89 & \textbf{91.58} & \textbf{77.04} & \textbf{93.30} & \textbf{81.42} & 78.82 & 57.42 \\
Safe-Net\textsuperscript{*} \cite{lin2024self62} \textcolor{blue}{[TIP’24]} & 91.22 & 86.06 &90.04  & 85.43  & 71.12  & 68.56  &  73.26 &  45.62 & 68.23  & 41.78  &  49.32 & 34.72  &  61.07 &  29.86 & 73.15  & 43.08  & 88.54  & 74.65  & 90.02  & 78.21  &  75.69 &  58.80 \\
CCR\textsuperscript{*} \cite{du2024ccr63} \textcolor{blue}{[TCSVT’24]} & 95.15& \textbf{91.80} & 90.93 & 80.62  & 81.83  & 73.89  &69.92   & 65.41  & 76.92  & 70.53  & 50.89  & 31.64  & 61.11  & 32.21  & 64.80  & 46.28  & 86.01  & 71.23  & 92.67  & 76.55  &   77.02&  64.02 \\
MuSe-Net\textsuperscript{*} \cite{wang2024Muse24} \textcolor{blue}{[PR’24]} & 88.02 & 75.10 & 87.87 & 69.85 & 87.73 & 71.12 & 83.74 & 66.52 &85.02 & 67.78 & 80.88 & 54.26 & 84.88 & 67.75 & 80.74 & 53.01 & 81.60 & 62.09 & 86.31 & 70.03 & 84.68 & 65.75 \\
\hline
\rowcolor{gray!15} 
Ours & 89.16 & 81.80 & 88.73 & 80.58 & \textbf{88.16} & \textbf{79.87} & \textbf{87.59} & \textbf{77.25} & \textbf{88.45} & \textbf{78.20} & \textbf{86.73} & \textbf{73.23} & \textbf{88.59} & \textbf{78.14} & \textbf{86.59} & \textbf{65.20} & 85.31 & 73.25 & 87.88 & 76.33 & \textbf{87.72} \textcolor{red}{($+3.04$)} & \textbf{76.39} \textcolor{red}{($+10.64$)} \\
\hline
\end{tabular}
}
\caption{\textbf{Performance (R@1(\%) and AP(\%)) on University-1652} for Drone $\rightarrow$ Satellite and Satellite $\rightarrow$ Drone tasks. In both tasks, drone-view images are stylized 10 different weather conditions, and the satellite-view images are constant. Best results are highlighted in bold. \textsuperscript{*} denotes the use of official pretrained weights.}
\label{tab1}
\vspace{-.1in}
\end{table*}

\begin{table*}[t]
\centering
\resizebox{\textwidth}{!}{%
\begin{tabular}{l|cc|cc|cc|cc|cc|cc|cc|cc|cc|cc|cc}
\hline
\multirow{2}{*}{Method} & \multicolumn{2}{c|}{Normal} & \multicolumn{2}{c|}{Fog} & \multicolumn{2}{c|}{Rain} & \multicolumn{2}{c|}{Snow} & \multicolumn{2}{c|}{Fog+Rain} & \multicolumn{2}{c|}{Fog+Snow} & \multicolumn{2}{c|}{Rain+Snow} & \multicolumn{2}{c|}{Dark} & \multicolumn{2}{c|}{Over-exp} & \multicolumn{2}{c|}{Wind} & \multicolumn{2}{c}{Mean} \\
 & R@1 & AP & R@1 & AP & R@1 & AP & R@1 & AP & R@1 & AP & R@1 & AP & R@1 & AP & R@1 & AP & R@1 & AP & R@1 & AP & R@1 & AP \\
\hline
\multicolumn{12}{r}{\textbf{Drone $\rightarrow$ Satellite}} \\
\hline
Zheng~\etal \cite{Zheng_Wei_Yang_202001} \textcolor{gray}{[backbone]} & 57.70 & 58.30 & 48.63 & 49.61 & 53.41 & 52.72 & 41.78 & 43.47 & 37.17 & 37.44 & 44.22 & 46.18 & 40.60 & 40.63 & 23.81 & 25.45 & 49.79 & 50.64 & 47.42 & 48.31 & 44.43 & 45.12 \\
IBN-Net \cite{Pan_Luo_Shi_Tang_201860} \textcolor{gray}{[backbone]} & 65.34 & 63.78 &56.03 & 56.57 & 55.73 & 58.55 & 47.80 & 49.53 & 43.45 & 44.98 & 50.04 & 51.00& 45.51& 45.92 & 29.61 & 30.93 & 56.01 & 56.96 & 57.36 & 58.10 & 50.69 & 51.63 \\
Sample4Geo\textsuperscript{\dag}  \cite{deuser2023sample4geo61} \textcolor{blue}{[ICCV’23]} & 74.93 & \textbf{78.76} & 72.58 & \textbf{76.44} & 34.60 & 41.56 & 28.95 &  35.02 & 35.10  & 41.47 & 12.95 & 17.90 & 20.05 & 25.95 & 34.18 & 38.99  & 38.40 & 43.68 & \textbf{67.80} &\textbf{72.41} &41.95 & 47.22 \\ 
Safe-Net\textsuperscript{*} \cite{lin2024self62} \textcolor{blue}{[TIP’24]} & 76.31 & 75.35 & \textbf{73.53} & 73.44  & 54.15  &  55.05 & 48.94  & 50.10  & 45.12  & 47.92  & 40.05  & 40.18  & 25.95  & 26.12  & 29.74  & 31.48  & 54.86  & 58.68  & 58.10  & 58.95  & 50.68  &  51.63 \\
CCR\textsuperscript{\dag}  \cite{du2024ccr63} \textcolor{blue}{[TCSVT’24]} & 73.22  & 74.53 & 70.95 & 73.14  &  60.14 & 64.95  &  50.31 &  53.12 & 45.87  & 49.14  &  45.80 &  47.87 & 31.25  & 32.94  & 31.03  & 34.36  & 59.97  & 61.07  & 52.02  &  53.33 & 52.06  & 53.46  \\
MuSe-Net\textsuperscript{*} \cite{wang2024Muse24} \textcolor{blue}{[PR’24]}  & 66.07 & 67.02 & 58.49 & 59.65& 58.94 & 60.14 & 54.85 & 56.12 & 44.31 & 45.82 & 49.81 & 51.26 & 49.42 & 50.87 & 29.34 & 31.03 & 55.02 & 56.36 & 59.97 & 61.05 & 52.02 & 53.33 \\
\hline
\rowcolor{gray!15} 
Ours & \textbf{76.72} & 75.51 & 68.49 & 68.87 & \textbf{71.77} & \textbf{71.20} & \textbf{59.95} & \textbf{60.62} & \textbf{58.24} & \textbf{58.83} & \textbf{64.36} & \textbf{66.27} & \textbf{58.49} & \textbf{58.89} & \textbf{40.42} & \textbf{55.75} & \textbf{61.57} & \textbf{71.70} & 65.19 & 67.00 & \textbf{62.52} \textcolor{red}{($+10.46$)} & \textbf{63.26} \textcolor{red}{($+9.80$)} \\
\hline
\multicolumn{12}{r}{\textbf{Satellite $\rightarrow$ Drone}} \\
\hline
Zheng~\etal \cite{Zheng_Wei_Yang_202001} \textcolor{gray}{[backbone]} & 70.20 & 57.98 & 63.77 & 46.90 & 68.72 & 50.85 & 61.72 & 39.70 & 62.10 & 32.75 & 71.70 & 40.39 & 59.72 & 37.55 & 45.49 & 25.28 & 52.11 & 43.40 & 56.62 & 45.31 & 61.21 & 42.01 \\
IBN-Net \cite{Pan_Luo_Shi_Tang_201860} \textcolor{gray}{[backbone]} & 73.68 & 62.91 & 67.41 & 55.75 & 72.30 & 56.44 & 64.07 & 47.69 & 66.98 & 39.54 & 71.10 & 47.32 & 68.46 & 45.95 & 54.72 & 31.53 & 65.64 & 53.77 & 73.48 & 57.03 & 67.79 & 49.79  \\
Sample4Geo\textsuperscript{\dag} \cite{deuser2023sample4geo61} \textcolor{blue}{[ICCV’23]} & 87.50 & 79.57 & 83.75 & 71.14 & 42.50 & 25.24 & 40.00 & 21.59  & 38.75 & 23.22 & 30.00 & 10.58 & 26.25 &  16.44  & 56.25 & 29.75 & 58.75 & 30.38 & \textbf{83.75} & \textbf{69.66} & 54.75 & 37.76 \\ 
Safe-Net\textsuperscript{*} \cite{lin2024self62} \textcolor{blue}{[TIP’24]} & 88.31 & 80.35 & 81.33 & 68.60  & 40.21  & 41.04  & 36.43  & 37.50  & 33.12  & 35.45  & 24.78  &  27.65 & 41.12  & 32.31  & 53.88  & 27.01  &  54.19 &  57.82 & 79.36  & 57.09  & 53.27  & 46.48  \\
CCR\textsuperscript{\dag}  \cite{du2024ccr63} \textcolor{blue}{[TCSVT’24]} & 90.59  & 80.45 & 82.99 & 70.62  & 43.39  & 45.90  & 39.81  & 40.88  & 42.63  & 39.46  & 29.32  & 30.65  &  25.89 & 26.94  & 26.01  & 30.40  & 58.01  &59.13   & 83.09  & 61.05  & 52.17  & 48.55  \\
MuSe-Net\textsuperscript{*} \cite{wang2024Muse24} \textcolor{blue}{[PR’24]} & 76.56 & 66.02 & 72.19 & 57.87 & 72.19 & 58.11 & 68.38 & 51.22 & 66.56 & 42.25 & 69.06 & 46.80 & 69.38 & 47.79 & 53.75 & 27.94 & 70.00 & 52.67 & 76.25 & 60.74 & 69.43 & 51.14 \\
\hline
\rowcolor{gray!15} 
Ours & \textbf{90.61} & \textbf{81.24} & \textbf{86.14} & \textbf{71.15} & \textbf{83.94} & \textbf{73.80} & \textbf{71.03} & \textbf{60.19} & \textbf{84.41} & \textbf{58.49} & \textbf{79.16} & \textbf{64.93} & \textbf{77.28} & \textbf{60.15} & \textbf{56.75} & \textbf{47.85} & \textbf{81.65} & \textbf{74.04} & 80.30 & 69.38 & \textbf{80.73} \textcolor{red}{($+11.30$)} & \textbf{66.12} \textcolor{red}{($+14.98$)} \\
\hline
\end{tabular}
}
\caption{\textbf{Performance (R@1(\%) and AP(\%)) on SUES-200} for Drone $\rightarrow$ Satellite and Satellite $\rightarrow$ Drone tasks. In both tasks, drone-view images are stylized in 10 weather conditions, while satellite-view images remain constant. Best results are highlighted in bold. \textsuperscript{*} denotes the use of official pretrained weights. \textsuperscript{\dag} denotes the use of official pretrained weights on University-1652.}
\label{tab2}
\end{table*}


\subsection{Comparison with Competitive Methods}
\begin{figure}[t]
    \centering
    \vspace{-.15in}
    \includegraphics[width=0.97\linewidth]{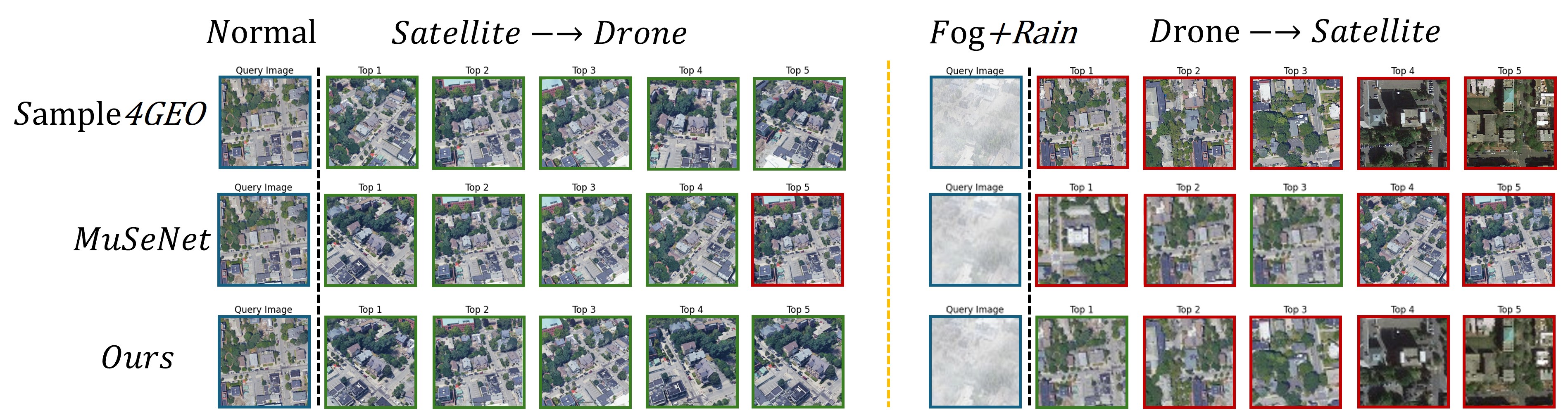}
    \caption{\textbf{Qualitative comparison under varying weather.} While existing methods perform reliably under clear weather, their accuracy drops markedly in adverse conditions. Our approach maintains superior localization performance, especially when drone images are severely affected by weather. Green boxes indicate correct matches, while images in red boxes represent incorrect matches.}
    \vspace{-.10in}
    \label{fig6}
\end{figure}

The experimental results on the University-1652 dataset are shown in Tab~\ref{tab1}.
We compare the proposed WeatherPrompt with several competitive cross-view geo-localization methods.
In the Drone \(\rightarrow\) Satellite task, WeatherPrompt achieves a mean Recall@1 (R@1) accuracy of 77.14\% (+11.99\%) and a mean Average Precision (AP) of 80.20\% (+11.04\%), outperforming existing state-of-the-art methods for multi-weather geo-localization. Similarly, in the Satellite \(\rightarrow\) Drone task, WeatherPrompt attains a mean R@1 accuracy of 87.72\% (+3.04\%) and a mean AP of 76.39\% (+10.64\%). 
Our method validates competitive performance when handling drone view images under different conditions, especially in terms of cross-view geo-localization in challenging weather scenarios (\eg, Dark, Fog+Snow, and Rain+Snow).
In addition to quantitative comparisons, we present qualitative retrieval results in Figure~\ref{fig6}. While existing methods can retrieve plausible matches in clear conditions, they suffer a substantial performance drop when encountering adverse weather conditions. In contrast, our approach consistently retrieves more accurate matches among the top-ranked candidates even in the presence of severe weather, highlighting its robustness and generalization capability.
In the real-world captured SUES-200 dataset, we combine drone-view images captured at four altitudes (150m, 200m, 250m, and 300m) to simulate diverse operational heights in practical drone applications, enabling a comprehensive evaluation of our method.
As shown in Tab~\ref{tab2}, compared to the state-of-the-art multi-weather geolocalization method MuSe-Net, our method shows significant improvements in several metrics. Specifically, Ours improves mean R@1 accuracy in the Drone \(\rightarrow\) Satellite task from 52.02\% to 61.20\% (+9.18\%) and mean AP from 53.33\% to 63.26\% (+9.80\%). In the Satellite \(\rightarrow\) Drone task, our method increases mean R@1 accuracy from 69.43\% to 80.73\% (+11.30\%) and mean AP from 51.14\% to 66.12\% (+14.98\%). 

\begin{table}[t]
  \centering
  \scriptsize
  \label{tab:4parts}

  \begin{subtable}[t]{0.95\textwidth}
    \centering
    \resizebox{0.95\textwidth}{!}{%
      \begin{tabular}{@{}lcccccccc@{}}
        \toprule
         & \multicolumn{4}{c}{D2S} 
         & \multicolumn{4}{c}{S2D} \\
        \cmidrule(lr){2-5}\cmidrule(l){6-9}
         & \multicolumn{2}{c}{Normal} & \multicolumn{2}{c}{Mean}
         & \multicolumn{2}{c}{Normal} & \multicolumn{2}{c}{Mean} \\
        \cmidrule(lr){2-3}\cmidrule(lr){4-5}
        \cmidrule(lr){6-7}\cmidrule(l){8-9}
        Method           & R@1(\%) & AP(\%) & R@1(\%) & AP(\%) 
                         & R@1(\%) & AP(\%) & R@1(\%) & AP(\%) \\
        \midrule
        +Concatenation          & 81.59 & 84.11 & 75.73 & 78.90  
                          & 88.59 & 81.76 & 85.38 &  73.84  \\
        +Static Gate      & 82.35 & 84.75 & 75.26 & 78.45   
                          & 88.45 & \textbf{82.25} & 85.89 & 74.61  \\
        \rowcolor{gray!15}
        +Dynamic Gate     & \textbf{82.78} & \textbf{85.18} & \textbf{77.14} & \textbf{80.20}   
                          & \textbf{89.16} & 81.80 & \textbf{87.72} & \textbf{76.39} \\
        \bottomrule
      \end{tabular}%
    }
    \caption{Impact of Weather-Driven Channel Gating on University-1652}\label{tab:gating_full}
  \end{subtable}


  \begin{subtable}[t]{0.48\textwidth}
    \centering
    \resizebox{\linewidth}{!}{%
    \begin{tabular}{ccccc}
      \toprule
      & \multicolumn{2}{c}{D2S} & \multicolumn{2}{c}{S2D} \\
      \cmidrule(lr){2-3}\cmidrule(lr){4-5}
      CoT Step & Mean R@1(\%) & Mean AP(\%) & Mean R@1(\%) & Mean AP(\%) \\
      \midrule
      NAN   & 74.53 & 78.72 & 85.17 & 73.26 \\
      0   & 75.10 & 79.15 & 85.45 & 73.65 \\ \hline 
      2   & 75.35 & 79.10 & 85.85 & 74.40  \\
      4   & 76.05 & 79.80 & 86.60 & 75.20  \\
      \rowcolor{gray!15}
      6   & \textbf{77.14} & \textbf{80.20} & \textbf{87.72} & \textbf{76.39}  \\
      \bottomrule
    \end{tabular}}
    \caption{Different CoT Setting on University-1652}\label{tab:cot_small}
  \end{subtable}
  \hfill
  \begin{subtable}[t]{0.48\linewidth}
    \centering
    \resizebox{\linewidth}{!}{%
    \begin{tabular}{clcccc}
    \hline
    \multirow{7}{*}{D2S}    & \multicolumn{5}{c}{Dark+Rain+Fog}   \\ \cline{2-6} 
       &  & R@1(\%)            & R@5(\%)    & R@10(\%)         & AP(\%)               \\ \hline
    &Sample4Geo~\cite{deuser2023sample4geo61}   &  22.22            & 61.11 & 77.78             &  31.65                     \\
    &MuSe-Net~\cite{wang2024Muse24}  &   22.22           & 66.67                &  83.33  & 31.70          \\ 
    \rowcolor{gray!15} 
    &Ours &  \textbf{44.44}              & \textbf{83.33}               &  \textbf{94.44}&\textbf{64.94}      \\ \hline       
    &Sample4Geo~\cite{deuser2023sample4geo61}   &  33.33             & 61.11            &  83.33 &43.47         \\
    S2D &MuSe-Net~\cite{wang2024Muse24}  & 38.89             & 61.11             &  88.89 & 44.34            \\ 
    \rowcolor{gray!15} 
    &Ours &  \textbf{66.66}              & \textbf{77.78}    &\textbf{94.44}           &  \textbf{72.15}           \\ \hline       
    \end{tabular}}
    \caption{Evaluation on Real World Videos}\label{tab:real}
  \end{subtable}
\caption{\textbf{Ablation Studies on University-1652 and Real-World Videos.} D2S denotes the Drone $\rightarrow$ Satellite and S2D denotes the Satellite $\rightarrow$ Drone.}
\vspace{-.30in}
\end{table}

\subsection{Ablation Studies and Further Discussion}

\textbf{Impact of Weather-Driven Gating Mechanism.}
We conduct ablation experiments to quantify the impact of the weather-driven gating module on multimodal fusion. Table~\ref{tab:gating_full} summarizes three alternative fusion strategies under identical settings: (1) Concatenation, which concatenates visual and textual features directly; (2) Static Gate, which fuses modalities using fixed average weights; and (3) Dynamic Gate (Ours), which adaptively reweights visual channels based on weather semantics. We find that the dynamic gating mechanism improves mean AP by 2.3\% over the no-gate baseline, with only 0.2M extra parameters and 0.4 ms added latency. This validates that weather-aware, channel-wise fusion enables more reliable feature integration, especially under adverse weather. The dynamic gate also reduces degradation from ambiguous weather descriptions, demonstrating its robustness for cross-weather geo-localization.

\textbf{Impact of Text-Guided Weather Semantics.}
We conduct an ablation study to assess how multi-weather textual supervision affects cross-view geo-localization.
Table~\ref{tab:cot_small} reports mean Recall@1 and mean Average Precision (AP) across two tasks (Drone $\rightarrow$ Satellite and Satellite $\rightarrow$ Drone) on University-1652, under varying levels of Chain-of-Thought (CoT) guidance.
The no-caption baseline (NAN) removes all text-based components, training the model with only the visual encoder and classification head.
When progressively increasing the CoT prompt steps, we observe consistent performance gains for both tasks.
Compared to the baseline, the best CoT setting (6-step) achieves improvements of +2.61\% mean Recall@1 and +1.48\% mean AP in the Drone $\rightarrow$ Satellite task, and +2.55\% mean Recall@1 and +3.13\% mean AP in the Satellite $\rightarrow$ Drone task.
Notably, all models leveraging text-based weather guidance achieve higher mean Recall@1 and AP compared to the visual-only baseline, demonstrating that even minimal textual supervision provides substantial benefits for model robustness and accuracy.

\textbf{Impact of Prompt Structuring on Multimodal Alignment.}
To comprehensively assess the impact of stepwise reasoning in prompt engineering on multimodal alignment, we generate captions using the same large vision–language model~\cite{Qwen2.5-VL12}, varying only the number of reasoning steps prescribed in the prompt. The zero-step baseline produces a single-turn weather description without explicit intermediate reasoning. In contrast, the Chain-of-Thought (CoT) variants employ multi-stage reasoning prompts, including 2-step, 4-step, and 6-step configurations.
The optimal 6-step CoT scheme decomposes the generation process into two reasoning chains: the first three steps sequentially estimate global visibility, local atmospheric cues, and  fine-grained weather semantics;      the latter three steps, conditioned on the inferred weather prior, successively infer macro scene layout, enumerate structural elements, and capture fine-grained spatial relationships. This structured decomposition compels the model to capture both meteorological variations and spatial semantics, yielding captions that are both semantically rich and consistent in format. As shown in Table~\ref{tab:cot_small}, model performance steadily improves as the number of reasoning steps increases, with the 6-step CoT achieving the best results in terms of both Recall@1 and mAP. These findings show that fine-grained, multi-stage reasoning significantly enhances cross-weather robustness and discriminative capacity, serving as a crucial ingredient for high-quality vision–language alignment in challenging multi-weather scenarios.


\textbf{Real-World Performance under Adverse Weather Conditions.} 
To validate real-world robustness, we collected 54 drone-satellite video pairs from YouTube to evaluate ours under three weather conditions, including Dark, Rain, and Fog.
As shown in Tab~\ref{tab:real}, our model consistently achieves superior results in terms of R@1, R@5, R@10, and Average Precision (AP). These results underscore the strong robustness and generalization capability of our model under challenging real-world scenarios with poor lighting or inclement weather conditions. 

\textbf{Limitations.} 
\label{limitations}
Despite of significant advancements in cross-view geo-localization under diverse weather conditions, several limitations inherited from external components warrant discussion: (1) Dataset. The evaluation relies on existing datasets, which lack exhaustive geographic and weather diversity. Their limited scope in representing globally rare or region-specific weather phenomena may affect generalization to unseen environmental extremes. 
(2) Language Model Biases. The weather and spatial captions generated by off-the-shelf vision-language models inherit biases from their pretraining corpora. Subtle inaccuracies in descriptor granularity, \eg, "haze" vs. "fog", could propagate into the alignment process.
Dataset and LVLM advances will mitigate these limitations.


\section{Conclusion}
In this paper, we propose a novel training-free, text-guided multi-modality alignment framework for robust cross-view geo-localization under complex and unseen weather conditions. By leveraging large vision–language models, we introduce a reasoning pipeline that automatically generates high-fidelity weather and spatial captions via chain-of-thought prompting, eliminating the need for costly manual descriptions or expert-controlled. Our multi-modal alignment model incorporates a dynamic channel-wise gating mechanism that adaptively fuses textual weather semantics with visual representations, achieving fine-grained disentanglement of scene and weather features. Extensive experiments on the University-1652 and SUES-200 benchmarks validate that our method consistently outperforms state-of-the-art approaches, particularly in challenging multi-weather scenarios.   Our approach sets a new paradigm for leveraging language-driven priors in aerial geo-localization and offers a scalable path toward real-world deployment under diverse environmental conditions. 

\section{Acknowledgement}
This work is supported by the Shanghai Committee of Science and Technology, China (Grant No.23ZR1423500), the National Natural Science Foundation of China under Grant No.62302287, University of Macau MYRG-GRG2024-00077-FST-UMDF and SRG2024-00002-FST, and the Science and Technology Development Fund (FDCT) 0043/2025/RIA1.

\printbibliography

@inproceedings{Zheng_Wei_Yang_202001,
  title        = {University-1652: A Multi-view Multi-source Benchmark for Drone-based Geo-localization},
  author       = {Zheng, Zhedong and Wei, Yunchao and Yang, Yi},
  booktitle    = {Proceedings of the 28th ACM International Conference on Multimedia (ACM MM)},
  pages        = {1395--1403},
  year         = {2020}
}

@article{Wang_Zheng_Yan_Zhang_Sun_Zheng_Yang_202202,
  title   = {Each Part Matters: Local Patterns Facilitate Cross-view Geo-localization},
  author  = {Wang, Tingyu and Zheng, Zhedong and Yan, Chenggang and Zhang, Jiyong and Sun, Yaoqi and Zheng, Bolun and Yang, Yi},
  journal = {IEEE Transactions on Circuits and Systems for Video Technology},
  year    = {2022},
  month   = feb,
  pages   = {867--879},
  doi     = {10.1109/TCSVT.2021.3061265}
}

@article{Lin_Zheng_Zhong_Luo_Li_Yang_Sebe_202203,
  title   = {Joint Representation Learning and Keypoint Detection for Cross-View Geo-Localization},
  author  = {Lin, Jinliang and Zheng, Zhedong and Zhong, Zhun and Luo, Zhiming and Li, Shaozi and Yang, Yi and Sebe, Nicu},
  journal = {IEEE Transactions on Image Processing},
  year    = {2022},
  pages   = {3780--3792},
  doi     = {10.1109/TIP.2022.3175601}
}

@article{zheng2017discriminatively,
  title   = {A Discriminatively Learned {CNN} Embedding for Person Re-identification},
  author  = {Zheng, Zhedong and Zheng, Liang and Yang, Yi},
  journal = {ACM Transactions on Multimedia Computing, Communications, and Applications (TOMM)},
  volume  = {14},
  number  = {1},
  pages   = {1--20},
  year    = {2017},
  doi     = {10.1145/3159171}
}

@inproceedings{feng2024multi04,
  title     = {Multi-weather Cross-view Geo-localization Using Denoising Diffusion Models},
  author    = {Feng, Tongtong and Li, Qing and Wang, Xin and Wang, Mingzi and Li, Guangyao and Zhu, Wenwu},
  booktitle = {Proceedings of the 2nd Workshop on UAVs in Multimedia: Capturing the World from a New Perspective},
  pages     = {35--39},
  year      = {2024}
}

@inproceedings{sindagi2020prior05,
  title     = {Prior-based Domain Adaptive Object Detection for Hazy and Rainy Conditions},
  author    = {Sindagi, Vishwanath A. and Oza, Poojan and Yasarla, Rajeev and Patel, Vishal M.},
  booktitle = {Proceedings of the European Conference on Computer Vision (ECCV)},
  year      = {2020}
}

@article{968495006,
  author  = {Zeng, Zelong and Wang, Zheng and Yang, Fan and Satoh, Shin’ichi},
  title   = {Geo-Localization via Ground-to-Satellite Cross-View Image Retrieval},
  journal = {IEEE Transactions on Multimedia},
  year    = {2023},
  volume  = {25},
  pages   = {2176--2188},
  doi     = {10.1109/TMM.2022.3144066}
}

@inproceedings{liu2022image07,
  title     = {Image-Adaptive {YOLO} for Object Detection in Adverse Weather Conditions},
  author    = {Liu, Wenyu and Ren, Gaofeng and Yu, Runsheng and Guo, Shi and Zhu, Jianke and Zhang, Lei},
  booktitle = {Proceedings of the AAAI Conference on Artificial Intelligence (AAAI)},
  volume    = {36},
  number    = {2},
  pages     = {1792--1800},
  year      = {2022}
}

@inproceedings{yang2023towards08,
  title     = {Towards Unified Text-based Person Retrieval: A Large-Scale Multi-Attribute and Language Search Benchmark},
  author    = {Yang, Shuyu and Zhou, Yinan and Zheng, Zhedong and Wang, Yaxiong and Zhu, Li and Wu, Yujiao},
  booktitle = {Proceedings of the 31st ACM International Conference on Multimedia (ACM MM)},
  pages     = {4492--4501},
  year      = {2023}
}

@inproceedings{yan2024dual09,
  title     = {Dual-Path Multimodal Optimal Transport for Composed Image Retrieval},
  author    = {Yan, Cairong and Ma, Meng and Zhang, Yanting and Wan, Yongquan},
  booktitle = {Proceedings of the Asian Conference on Computer Vision (ACCV)},
  pages     = {1741--1755},
  year      = {2024}
}

@article{yan2023clip10,
  title   = {{CLIP}-Driven Fine-Grained Text-Image Person Re-identification},
  author  = {Yan, Shuanglin and Dong, Neng and Zhang, Liyan and Tang, Jinhui},
  journal = {IEEE Transactions on Image Processing},
  volume  = {32},
  pages   = {6032--6046},
  year    = {2023}
}

@inproceedings{vendrow2024inquire11,
  title     = {{INQUIRE}: A Natural World Text-to-Image Retrieval Benchmark},
  author    = {Vendrow, Edward and Pantazis, Omiros and Shepard, Alexander and Brostow, Gabriel and Jones, Kate and Mac Aodha, Oisin and Beery, Sara and Van Horn, Grant},
  booktitle = {Advances in Neural Information Processing Systems (NeurIPS)},
  volume    = {37},
  pages     = {126500--126514},
  year      = {2024}
}

@article{Qwen2.5-VL12,
  title   = {Qwen2.5-{VL} Technical Report},
  author  = {Bai, Shuai and Chen, Keqin and Liu, Xuejing and Wang, Jialin and Ge, Wenbin and Song, Sibo and Dang, Kai and Wang, Peng and Wang, Shijie and Tang, Jun and Zhong, Humen and Zhu, Yuanzhi and Yang, Mingkun and Li, Zhaohai and Wan, Jianqiang and Wang, Pengfei and Ding, Wei and Fu, Zheren and Xu, Yiheng and Ye, Jiabo and Zhang, Xi and Xie, Tianbao and Cheng, Zesen and Zhang, Hang and Yang, Zhibo and Xu, Haiyang and Lin, Junyang},
  journal = {arXiv preprint},
  year    = {2025},
  eprint  = {2502.13923},
  archivePrefix = {arXiv},
  primaryClass  = {cs.CV}
}

@inproceedings{wei2022chain13,
  title     = {Chain-of-Thought Prompting Elicits Reasoning in Large Language Models},
  author    = {Wei, Jason and Wang, Xuezhi and Schuurmans, Dale and Bosma, Maarten and Xia, Fei and Chi, Ed and Le, Quoc V. and Zhou, Denny and others},
  booktitle = {Advances in Neural Information Processing Systems (NeurIPS)},
  volume    = {35},
  pages     = {24824--24837},
  year      = {2022}
}

@inproceedings{berton2022deep14,
  title     = {Deep Visual Geo-localization Benchmark},
  author    = {Berton, Gabriele and Mereu, Riccardo and Trivigno, Gabriele and Masone, Carlo and Csurka, Gabriela and Sattler, Torsten and Caputo, Barbara},
  booktitle = {Proceedings of the IEEE/CVF Conference on Computer Vision and Pattern Recognition (CVPR)},
  pages     = {5396--5407},
  year      = {2022}
}

@inproceedings{jin2017learned15,
  title     = {Learned Contextual Feature Reweighting for Image Geo-localization},
  author    = {Kim, Hyo Jin and Dunn, Enrique and Frahm, Jan-Michael},
  booktitle = {Proceedings of the IEEE Conference on Computer Vision and Pattern Recognition (CVPR)},
  pages     = {2136--2145},
  year      = {2017}
}

@inproceedings{chu2024towards16,
  title     = {Towards Natural Language-Guided Drones: {GeoText-1652} Benchmark with Spatial Relation Matching},
  author    = {Chu, Meng and Zheng, Zhedong and Ji, Wei and Wang, Tingyu and Chua, Tat-Seng},
  booktitle = {Proceedings of the European Conference on Computer Vision (ECCV)},
  pages     = {213--231},
  year      = {2024}
}

@article{cruz2012scale17,
  title   = {Scale Invariant Feature Transform on the Sphere: Theory and Applications},
  author  = {Cruz-Mota, Javier and Bogdanova, Iva and Paquier, Beno{\^\i}t and Bierlaire, Michel and Thiran, Jean-Philippe},
  journal = {International Journal of Computer Vision},
  volume  = {98},
  pages   = {217--241},
  year    = {2012}
}

@inproceedings{bay2006surf18,
  title     = {{SURF}: Speeded Up Robust Features},
  author    = {Bay, Herbert and Tuytelaars, Tinne and Van Gool, Luc},
  booktitle = {Proceedings of the European Conference on Computer Vision (ECCV)},
  year      = {2006}
}

@article{zhang2021vector19,
  title   = {Vector of Locally and Adaptively Aggregated Descriptors for Image Feature Representation},
  author  = {Zhang, Jian and Cao, Yunyin and Wu, Qun},
  journal = {Pattern Recognition},
  volume  = {116},
  pages   = {107952},
  year    = {2021}
}

@article{sanchez2013image20,
  title   = {Image Classification with the Fisher Vector: Theory and Practice},
  author  = {S{\'a}nchez, Jorge and Perronnin, Florent and Mensink, Thomas and Verbeek, Jakob},
  journal = {International Journal of Computer Vision},
  volume  = {105},
  pages   = {222--245},
  year    = {2013}
}

@inproceedings{barath2022learning21,
  title     = {Learning to Find Good Models in {RANSAC}},
  author    = {Barath, Daniel and Cavalli, Luca and Pollefeys, Marc},
  booktitle = {Proceedings of the IEEE/CVF Conference on Computer Vision and Pattern Recognition (CVPR)},
  pages     = {15744--15753},
  year      = {2022}
}

@inproceedings{Castaldo_Zamir_Angst_Palmieri_Savarese_201522,
  title     = {Semantic Cross-View Matching},
  author    = {Castaldo, Francesco and Zamir, Amir and Angst, Roland and Palmieri, Francesco and Savarese, Silvio},
  booktitle = {2015 IEEE International Conference on Computer Vision Workshops (ICCVW)},
  year      = {2015},
  doi       = {10.1109/ICCVW.2015.137}
}

@inproceedings{Lin_Belongie_Hays_201323,
  title     = {Cross-View Image Geolocalization},
  author    = {Lin, Tsung-Yi and Belongie, Serge and Hays, James},
  booktitle = {Proceedings of the IEEE Conference on Computer Vision and Pattern Recognition (CVPR)},
  year      = {2013},
  doi       = {10.1109/CVPR.2013.120}
}

@article{wang2024Muse24,
  title   = {Multiple-environment Self-adaptive Network for Aerial-view Geo-localization},
  author  = {Wang, Tingyu and Zheng, Zhedong and Sun, Yaoqi and Yan, Chenggang and Yang, Yi and Chua, Tat-Seng},
  journal = {Pattern Recognition},
  volume  = {152},
  pages   = {110363},
  year    = {2024},
  doi     = {10.1016/j.patcog.2024.110363}
}

@inproceedings{Workman_Souvenir_Jacobs_201525,
  title     = {Wide-Area Image Geolocalization with Aerial Reference Imagery},
  author    = {Workman, Scott and Souvenir, Richard and Jacobs, Nathan},
  booktitle = {Proceedings of the IEEE International Conference on Computer Vision (ICCV)},
  year      = {2015}
}

@inproceedings{Lin_YinCui_Belongie_Hays_201526,
  title     = {Learning Deep Representations for Ground-to-Aerial Geolocalization},
  author    = {Lin, Tsung-Yi and Cui, Yin and Belongie, Serge and Hays, James},
  booktitle = {Proceedings of the IEEE Conference on Computer Vision and Pattern Recognition (CVPR)},
  year      = {2015}
}

@inproceedings{yang2021cross27,
  title     = {Cross-View Geo-localization with Layer-to-Layer Transformer},
  author    = {Yang, Hongji and Lu, Xiufan and Zhu, Yingying},
  booktitle = {Advances in Neural Information Processing Systems (NeurIPS)},
  volume    = {34},
  pages     = {29009--29020},
  year      = {2021}
}

@inproceedings{rodrigues2022global28,
  title     = {Global Assists Local: Effective Aerial Representations for Field-of-View Constrained Image Geo-localization},
  author    = {Rodrigues, Royston and Tani, Masahiro},
  booktitle = {Proceedings of the IEEE/CVF Winter Conference on Applications of Computer Vision (WACV)},
  pages     = {3871--3879},
  year      = {2022}
}

@article{dai2021transformer29,
  title   = {A Transformer-based Feature Segmentation and Region Alignment Method for {UAV}-view Geo-localization},
  author  = {Dai, Ming and Hu, Jianhong and Zhuang, Jiedong and Zheng, Enhui},
  journal = {IEEE Transactions on Circuits and Systems for Video Technology},
  volume  = {32},
  number  = {7},
  pages   = {4376--4389},
  year    = {2021}
}

@inproceedings{ma2022both30,
  title     = {Both Style and Fog Matter: Cumulative Domain Adaptation for Semantic Foggy Scene Understanding},
  author    = {Ma, Xianzheng and Wang, Zhixiang and Zhan, Yacheng and Zheng, Yinqiang and Wang, Zheng and Dai, Dengxin and Lin, Chia-Wen},
  booktitle = {Proceedings of the IEEE/CVF Conference on Computer Vision and Pattern Recognition (CVPR)},
  pages     = {18922--18931},
  year      = {2022}
}

@article{tan2023mapd31,
  title   = {{MAPD}: An {FPGA}-based Real-time Video Haze Removal Accelerator Using Mixed Atmosphere Prior},
  author  = {Tan, Yanjie and Zhu, Yifu and Huang, Zhaoyang and Tan, Huailiang and Li, Keqin},
  journal = {IEEE Transactions on Computer-Aided Design of Integrated Circuits and Systems},
  volume  = {42},
  number  = {12},
  pages   = {4777--4790},
  year    = {2023}
}

@inproceedings{zhang2023cross32,
  title     = {Cross-View Geo-localization via Learning Disentangled Geometric Layout Correspondence},
  author    = {Zhang, Xiaohan and Li, Xingyu and Sultani, Waqas and Zhou, Yi and Wshah, Safwan},
  booktitle = {Proceedings of the AAAI Conference on Artificial Intelligence (AAAI)},
  volume    = {37},
  number    = {3},
  pages     = {3480--3488},
  year      = {2023}
}

@inproceedings{zhu2022transgeo33,
  title     = {TransGeo: Transformer is All You Need for Cross-View Image Geo-localization},
  author    = {Zhu, Sijie and Shah, Mubarak and Chen, Chen},
  booktitle = {Proceedings of the IEEE/CVF Conference on Computer Vision and Pattern Recognition (CVPR)},
  pages     = {1162--1171},
  year      = {2022}
}

@inproceedings{radford2021learning34,
  title     = {Learning Transferable Visual Models from Natural Language Supervision},
  author    = {Radford, Alec and Kim, Jong Wook and Hallacy, Chris and Ramesh, Aditya and Goh, Gabriel and Agarwal, Sandhini and Sastry, Girish and Askell, Amanda and Mishkin, Pamela and Clark, Jack and others},
  booktitle = {Proceedings of the 38th International Conference on Machine Learning (ICML)},
  series    = {Proceedings of Machine Learning Research},
  volume    = {139},
  pages     = {8748--8763},
  year      = {2021},
  publisher = {PMLR}
}

@inproceedings{li2022blip35,
  title     = {{BLIP}: Bootstrapping Language-Image Pre-training for Unified Vision-Language Understanding and Generation},
  author    = {Li, Junnan and Li, Dongxu and Xiong, Caiming and Hoi, Steven C. H.},
  booktitle = {Proceedings of the 39th International Conference on Machine Learning (ICML)},
  series    = {Proceedings of Machine Learning Research},
  volume    = {162},
  pages     = {12888--12900},
  year      = {2022},
  publisher = {PMLR}
}

@article{xvlm36,
  title   = {Multi-Grained Vision Language Pre-Training: Aligning Texts with Visual Concepts},
  author  = {Zeng, Yan and Zhang, Xinsong and Li, Hang},
  journal = {arXiv preprint},
  year    = {2021},
  eprint  = {2111.08276},
  archivePrefix = {arXiv},
  primaryClass  = {cs.CV}
}

@article{zhu2024mvp37,
  title   = {{MVP}: Meta Visual Prompt Tuning for Few-shot Remote Sensing Image Scene Classification},
  author  = {Zhu, Junjie and Li, Yiying and Yang, Ke and Guan, Naiyang and Fan, Zunlin and Qiu, Chunping and Yi, Xiaodong},
  journal = {IEEE Transactions on Geoscience and Remote Sensing},
  volume  = {62},
  pages   = {1--13},
  year    = {2024}
}

@article{chen2024rsprompter38,
  title   = {{RSPrompter}: Learning to Prompt for Remote Sensing Instance Segmentation Based on Visual Foundation Model},
  author  = {Chen, Keyan and Liu, Chenyang and Chen, Hao and Zhang, Haotian and Li, Wenyuan and Zou, Zhengxia and Shi, Zhenwei},
  journal = {IEEE Transactions on Geoscience and Remote Sensing},
  volume  = {62},
  pages   = {1--17},
  year    = {2024}
}

@article{yuan2023parameter39,
  title   = {Parameter-efficient Transfer Learning for Remote Sensing Image--Text Retrieval},
  author  = {Yuan, Yuan and Zhan, Yang and Xiong, Zhitong},
  journal = {IEEE Transactions on Geoscience and Remote Sensing},
  volume  = {61},
  pages   = {1--14},
  year    = {2023}
}

@article{zheng2020dual40,
  title   = {Dual-Path Convolutional Image-Text Embeddings with Instance Loss},
  author  = {Zheng, Zhedong and Zheng, Liang and Garrett, Michael and Yang, Yi and Xu, Mingliang and Shen, Yi-Dong},
  journal = {ACM Transactions on Multimedia Computing, Communications, and Applications (TOMM)},
  volume  = {16},
  number  = {2},
  pages   = {1--23},
  year    = {2020}
}

@inproceedings{wang2019camp41,
  title     = {{CAMP}: Cross-Modal Adaptive Message Passing for Text-Image Retrieval},
  author    = {Wang, Zihao and Liu, Xihui and Li, Hongsheng and Sheng, Lu and Yan, Junjie and Wang, Xiaogang and Shao, Jing},
  booktitle = {Proceedings of the IEEE/CVF International Conference on Computer Vision (ICCV)},
  pages     = {5764--5773},
  year      = {2019}
}

@inproceedings{chen2020uniter43,
  title     = {{UNITER}: Universal Image-Text Representation Learning},
  author    = {Chen, Yen-Chun and Li, Linjie and Yu, Licheng and El Kholy, Ahmed and Ahmed, Faisal and Gan, Zhe and Cheng, Yu and Liu, Jingjing},
  booktitle = {Proceedings of the European Conference on Computer Vision (ECCV)},
  pages     = {104--120},
  year      = {2020}
}

@inproceedings{li2020oscar44,
  title     = {{OSCAR}: Object-Semantics Aligned Pre-training for Vision-Language Tasks},
  author    = {Li, Xiujun and Yin, Xi and Li, Chunyuan and Zhang, Pengchuan and Hu, Xiaowei and Zhang, Lei and Wang, Lijuan and Hu, Houdong and Dong, Li and Wei, Furu and others},
  booktitle = {Proceedings of the European Conference on Computer Vision (ECCV)},
  pages     = {121--137},
  year      = {2020}
}

@inproceedings{brown2020language45,
  title     = {Language Models are Few-Shot Learners},
  author    = {Brown, Tom and Mann, Benjamin and Ryder, Nick and Subbiah, Melanie and Kaplan, Jared D. and Dhariwal, Prafulla and Neelakantan, Arvind and Shyam, Pranav and Sastry, Girish and Askell, Amanda and others},
  booktitle = {Advances in Neural Information Processing Systems (NeurIPS)},
  volume    = {33},
  pages     = {1877--1901},
  year      = {2020}
}

@article{achiam2023gpt46,
  title   = {{GPT-4} Technical Report},
  author  = {Achiam, Josh and Adler, Steven and Agarwal, Sandhini and others},
  journal = {arXiv preprint},
  year    = {2023},
  eprint  = {2303.08774},
  archivePrefix = {arXiv},
  primaryClass  = {cs.CL}
}

@inproceedings{chen2022visualgpt47,
  title     = {VisualGPT: Data-efficient Adaptation of Pretrained Language Models for Image Captioning},
  author    = {Chen, Jun and Guo, Han and Yi, Kai and Li, Boyang and Elhoseiny, Mohamed},
  booktitle = {Proceedings of the IEEE/CVF Conference on Computer Vision and Pattern Recognition (CVPR)},
  pages     = {18030--18040},
  year      = {2022}
}

@article{zhang2023multimodal48,
  title   = {Multimodal Chain-of-Thought Reasoning in Language Models},
  author  = {Zhang, Zhuosheng and Zhang, Aston and Li, Mu and Zhao, Hai and Karypis, George and Smola, Alex},
  journal = {arXiv preprint},
  year    = {2023},
  eprint  = {2302.00923},
  archivePrefix = {arXiv},
  primaryClass  = {cs.CL}
}

@inproceedings{sriramanan2024llm49,
  title     = {{LLM}-Check: Investigating Detection of Hallucinations in Large Language Models},
  author    = {Sriramanan, Gaurang and Bharti, Siddhant and Sadasivan, Vinu Sankar and Saha, Shoumik and Kattakinda, Priyatham and Feizi, Soheil},
  booktitle = {Advances in Neural Information Processing Systems (NeurIPS)},
  volume    = {37},
  pages     = {34188--34216},
  year      = {2024}
}

@inproceedings{jiang2024hallucination50,
  title     = {Hallucination Augmented Contrastive Learning for Multimodal Large Language Model},
  author    = {Jiang, Chaoya and Xu, Haiyang and Dong, Mengfan and Chen, Jiaxing and Ye, Wei and Yan, Ming and Ye, Qinghao and Zhang, Ji and Huang, Fei and Zhang, Shikun},
  booktitle = {Proceedings of the IEEE/CVF Conference on Computer Vision and Pattern Recognition (CVPR)},
  pages     = {27036--27046},
  year      = {2024}
}

@inproceedings{ju2024video2bev,
  title     = {Video2BEV: Transforming Drone Videos to {BEVs} for Video-based Geo-localization},
  author    = {Ju, Hao and Huang, Shaofei and Liu, Si and Zheng, Zhedong},
  booktitle = {Proceedings of the IEEE/CVF International Conference on Computer Vision (ICCV)},
  year      = {2024}
}

@inproceedings{kim2024exploiting51,
  title     = {Exploiting Semantic Reconstruction to Mitigate Hallucinations in Vision-Language Models},
  author    = {Kim, Minchan and Kim, Minyeong and Bae, Junik and Choi, Suhwan and Kim, Sungkyung and Chang, Buru},
  booktitle = {Proceedings of the European Conference on Computer Vision (ECCV)},
  pages     = {236--252},
  year      = {2024}
}

@article{ye2024cross52,
  title   = {Where am I? Cross-View Geo-localization with Natural Language Descriptions},
  author  = {Ye, Junyan and Lin, Honglin and Ou, Leyan and Chen, Dairong and Wang, Zihao and Zhu, Qi and He, Conghui and Li, Weijia},
  journal = {arXiv preprint},
  year    = {2024},
  eprint  = {2412.17007},
  archivePrefix = {arXiv},
  primaryClass  = {cs.CV}
}

@article{robbins1951stochastic53,
  title   = {A Stochastic Approximation Method},
  author  = {Robbins, Herbert and Monro, Sutton},
  journal = {The Annals of Mathematical Statistics},
  pages   = {400--407},
  year    = {1951}
}

@inproceedings{devlin2019bert54,
  title     = {{BERT}: Pre-training of Deep Bidirectional Transformers for Language Understanding},
  author    = {Devlin, Jacob and Chang, Ming-Wei and Lee, Kenton and Toutanova, Kristina},
  booktitle = {Proceedings of the 2019 Conference of the North American Chapter of the Association for Computational Linguistics (NAACL-HLT)},
  pages     = {4171--4186},
  year      = {2019}
}

@inproceedings{liu2021swin55,
  title     = {Swin Transformer: Hierarchical Vision Transformer Using Shifted Windows},
  author    = {Liu, Ze and Lin, Yutong and Cao, Yue and Hu, Han and Wei, Yixuan and Zhang, Zheng and Lin, Stephen and Guo, Baining},
  booktitle = {Proceedings of the IEEE/CVF International Conference on Computer Vision (ICCV)},
  pages     = {10012--10022},
  year      = {2021}
}

@inproceedings{paszke2019pytorch56,
  title     = {{PyTorch}: An Imperative Style, High-Performance Deep Learning Library},
  author    = {Paszke, Adam and Gross, Sam and Massa, Francisco and Lerer, Adam and Bradbury, James and Chanan, Gregory and Killeen, Trevor and Lin, Zeming and Gimelshein, Natalia and Antiga, Luca and others},
  booktitle = {Advances in Neural Information Processing Systems (NeurIPS)},
  volume    = {32},
  year      = {2019}
}

@article{zhu2023sues57,
  title   = {{SUES-200}: A Multi-height Multi-scene Cross-view Image Benchmark Across Drone and Satellite},
  author  = {Zhu, Runzhe and Yin, Ling and Yang, Mingze and Wu, Fei and Yang, Yuncheng and Hu, Wenbo},
  journal = {IEEE Transactions on Circuits and Systems for Video Technology},
  volume  = {33},
  number  = {9},
  pages   = {4825--4839},
  year    = {2023}
}

@inproceedings{He_Zhang_Ren_Sun_201658,
  title     = {Deep Residual Learning for Image Recognition},
  author    = {He, Kaiming and Zhang, Xiangyu and Ren, Shaoqing and Sun, Jian},
  booktitle = {Proceedings of the IEEE Conference on Computer Vision and Pattern Recognition (CVPR)},
  year      = {2016},
  doi       = {10.1109/CVPR.2016.90}
}

@inproceedings{huang2017densely59,
  title     = {Densely Connected Convolutional Networks},
  author    = {Huang, Gao and Liu, Zhuang and Van Der Maaten, Laurens and Weinberger, Kilian Q.},
  booktitle = {Proceedings of the IEEE Conference on Computer Vision and Pattern Recognition (CVPR)},
  pages     = {4700--4808},
  year      = {2017}
}

@inproceedings{Pan_Luo_Shi_Tang_201860,
  author    = {Pan, Xingang and Luo, Ping and Shi, Jianping and Tang, Xiaoou},
  title     = {Two at Once: Enhancing Learning and Generalization Capacities via {IBN}-Net},
  booktitle = {Proceedings of the European Conference on Computer Vision (ECCV)},
  pages     = {484--500},
  year      = {2018}
}

@inproceedings{deuser2023sample4geo61,
  title     = {Sample4Geo: Hard Negative Sampling for Cross-View Geo-Localization},
  author    = {Deuser, Fabian and Habel, Konrad and Oswald, Norbert},
  booktitle = {Proceedings of the IEEE/CVF International Conference on Computer Vision (ICCV)},
  pages     = {16847--16856},
  year      = {2023}
}

@article{lin2024self62,
  title   = {A Self-Adaptive Feature Extraction Method for Aerial-View Geo-Localization},
  author  = {Lin, Jinliang and Luo, Zhiming and Lin, Dazhen and Li, Shaozi and Zhong, Zhun},
  journal = {IEEE Transactions on Image Processing},
  year    = {2024}
}

@article{du2024ccr63,
  title   = {{CCR}: A Counterfactual Causal Reasoning-Based Method for Cross-View Geo-Localization},
  author  = {Du, Haolin and He, Jingfei and Zhao, Yuanqing},
  journal = {IEEE Transactions on Circuits and Systems for Video Technology},
  year    = {2024}
}

@inproceedings{li2022fine64,
  title     = {Fine-grained Semantically Aligned Vision-Language Pre-training},
  author    = {Li, Juncheng and He, Xin and Wei, Longhui and Qian, Long and Zhu, Linchao and Xie, Lingxi and Zhuang, Yueting and Tian, Qi and Tang, Siliang},
  booktitle = {Advances in Neural Information Processing Systems (NeurIPS)},
  volume    = {35},
  pages     = {7290--7303},
  year      = {2022}
}

@article{chow2024unified65,
  title   = {Unified Generative and Discriminative Training for Multi-modal Large Language Models},
  author  = {Chow, Wei and Li, Juncheng and Yu, Qifan and Pan, Kaihang and Fei, Hao and Ge, Zhiqi and Yang, Shuai and Tang, Siliang and Zhang, Hanwang and Sun, Qianru},
  journal = {arXiv preprint},
  year    = {2024},
  eprint  = {2411.00304},
  archivePrefix = {arXiv},
  primaryClass  = {cs.CV}
}

@inproceedings{vepa2024integrating66,
  title     = {Integrating Deep Metric Learning with Coreset for Active Learning in 3D Segmentation},
  author    = {Vepa, Arvind M. and Yang, Zukang and Choi, Andrew and Joo, Jungseock and Scalzo, Fabien and Sun, Yizhou},
  booktitle = {Advances in Neural Information Processing Systems (NeurIPS)},
  volume    = {37},
  pages     = {71643--71671},
  year      = {2024}
}

@misc{imgaug67,
  author       = {Jung, Alexander B. and Wada, Kentaro and Crall, Jon and Tanaka, Satoshi and Graving, Jake and Reinders, Christoph and Yadav, Sarthak and Banerjee, Joy and Vecsei, G{\'a}bor and Kraft, Adam and Rui, Zheng and Borovec, Jirka and Vallentin, Christian and Zhydenko, Semen and Pfeiffer, Kilian and Cook, Ben and Fern{\'a}ndez, Ismael and De Rainville, Fran{\c{c}}ois-Michel and Weng, Chi-Hung and Ayala-Acevedo, Abner and Meudec, Raphael and Laporte, Matias and others},
  title        = {{imgaug}},
  howpublished = {\url{https://github.com/aleju/imgaug}},
  note         = {Accessed: 2020-02-01},
  year         = {2020}
}

@inproceedings{choi2023depth68,
  title     = {Depth-Discriminative Metric Learning for Monocular 3D Object Detection},
  author    = {Choi, Wonhyeok and Shin, Mingyu and Im, Sunghoon},
  booktitle = {Advances in Neural Information Processing Systems (NeurIPS)},
  volume    = {36},
  pages     = {80165--80177},
  year      = {2023}
}

@article{qu2024lush69,
  title   = {LuSh-{NeRF}: Lighting Up and Sharpening {NeRFs} for Low-light Scenes},
  author  = {Qu, Zefan and Xu, Ke and Hancke, Gerhard Petrus and Lau, Rynson W. H.},
  journal = {arXiv preprint},
  year    = {2024},
  eprint  = {2411.06757},
  archivePrefix = {arXiv},
  primaryClass  = {cs.CV}
}

@article{chen2024restoreagent70,
  title   = {RestoreAgent: Autonomous Image Restoration Agent via Multimodal Large Language Models},
  author  = {Chen, Haoyu and Li, Wenbo and Gu, Jinjin and Ren, Jingjing and Chen, Sixiang and Ye, Tian and Pei, Renjing and Zhou, Kaiwen and Song, Fenglong and Zhu, Lei},
  journal = {arXiv preprint},
  year    = {2024},
  eprint  = {2407.18035},
  archivePrefix = {arXiv},
  primaryClass  = {cs.CV}
}

@inproceedings{zhu2022demo87,
  title     = {Demo Abstract: An {UAV}-based 3D Spectrum Real-time Mapping System},
  author    = {Zhu, Qiuming and Zhao, Yi and Huang, Yang and Lin, Zhipeng and Wang, Lu HanJie and Bai, Yunpeng and Lan, Tianxu and Zhou, Fuhui and Wu, Qihui},
  booktitle = {IEEE INFOCOM 2022 -- IEEE Conference on Computer Communications Workshops (INFOCOM WKSHPS)},
  pages     = {1--2},
  year      = {2022}
}

@article{wang2024sparse88,
  title   = {Sparse Bayesian Learning-based Hierarchical Construction for 3D Radio Environment Maps Incorporating Channel Shadowing},
  author  = {Wang, Jie and Zhu, Qiuming and Lin, Zhipeng and Chen, Junting and Ding, Guoru and Wu, Qihui and Gu, Guochen and Gao, Qianhao},
  journal = {IEEE Transactions on Wireless Communications},
  year    = {2024}
}

\newpage





\clearpage

\newpage
\section*{NeurIPS Paper Checklist}

\begin{enumerate}

\item {\bf Claims}
    \item[] Question: Do the main claims made in the abstract and introduction accurately reflect the paper's contributions and scope?
    \item[] Answer: \answerYes{} 
    \item[] Justification: The main claims in the abstract and introduction accurately reflect the paper's core contributions and scope. The paper proposes a training-free weather reasoning approach for multi-weather geo-localization, introduces a text-guided feature alignment framework with dynamic gating, and yields consistent improvements on challenging benchmarks. All major claims are substantiated by method descriptions and experimental results throughout the paper.

\item {\bf Limitations}
    \item[] Question: Does the paper discuss the limitations of the work performed by the authors?
    \item[] Answer: \answerYes{} 
    \item[] Justification: The limitations of the work are discussed in Section~\ref{limitations}.
    \item[] Guidelines:
    \begin{itemize}
        \item The answer NA means that the paper has no limitation while the answer No means that the paper has limitations, but those are not discussed in the paper. 
        \item The authors are encouraged to create a separate "Limitations" section in their paper.
        \item The paper should point out any strong assumptions and how robust the results are to violations of these assumptions (\eg, independence assumptions, noiseless settings, model well-specification, asymptotic approximations only holding locally). The authors should reflect on how these assumptions might be violated in practice and what the implications would be.
        \item The authors should reflect on the scope of the claims made, \eg, if the approach was only tested on a few datasets or with a few runs. In general, empirical results often depend on implicit assumptions, which should be articulated.
        \item The authors should reflect on the factors that influence the performance of the approach. For example, a facial recognition algorithm may perform poorly when image resolution is low or images are taken in low lighting. Or a speech-to-text system might not be used reliably to provide closed captions for online lectures because it fails to handle technical jargon.
        \item The authors should discuss the computational efficiency of the proposed algorithms and how they scale with dataset size.
        \item If applicable, the authors should discuss possible limitations of their approach to address problems of privacy and fairness.
        \item While the authors might fear that complete honesty about limitations might be used by reviewers as grounds for rejection, a worse outcome might be that reviewers discover limitations that aren't acknowledged in the paper. The authors should use their best judgment and recognize that individual actions in favor of transparency play an important role in developing norms that preserve the integrity of the community. Reviewers will be specifically instructed to not penalize honesty concerning limitations.
    \end{itemize}

\item {\bf Theory assumptions and proofs}
    \item[] Question: For each theoretical result, does the paper provide the full set of assumptions and a complete (and correct) proof?
    \item[] Answer: \answerNo{}
    \item[] Justification: The paper does not present formal theorems or proofs.

    \item {\bf Experimental result reproducibility}
    \item[] Question: Does the paper fully disclose all the information needed to reproduce the main experimental results of the paper to the extent that it affects the main claims and/or conclusions of the paper (regardless of whether the code and data are provided or not)?
    \item[] Answer: \answerYes{} 
    \item[] Justification: The paper provides all necessary details for experimental reproducibility. All modules, loss functions, and augmentation strategies are specified in the main text. To further support reproducibility, we will release our code and configuration files upon publication.
    \item[] Guidelines:
    \begin{itemize}
        \item The answer NA means that the paper does not include experiments.
        \item If the paper includes experiments, a No answer to this question will not be perceived well by the reviewers: Making the paper reproducible is important, regardless of whether the code and data are provided or not.
        \item If the contribution is a dataset and/or model, the authors should describe the steps taken to make their results reproducible or verifiable. 
        \item Depending on the contribution, reproducibility can be accomplished in various ways. For example, if the contribution is a novel architecture, describing the architecture fully might suffice, or if the contribution is a specific model and empirical evaluation, it may be necessary to either make it possible for others to replicate the model with the same dataset, or provide access to the model. In general. releasing code and data is often one good way to accomplish this, but reproducibility can also be provided via detailed instructions for how to replicate the results, access to a hosted model (\eg, in the case of a large language model), releasing of a model checkpoint, or other means that are appropriate to the research performed.
        \item While NeurIPS does not require releasing code, the conference does require all submissions to provide some reasonable avenue for reproducibility, which may depend on the nature of the contribution. For example
        \begin{enumerate}
            \item If the contribution is primarily a new algorithm, the paper should make it clear how to reproduce that algorithm.
            \item If the contribution is primarily a new model architecture, the paper should describe the architecture clearly and fully.
            \item If the contribution is a new model (\eg, a large language model), then there should either be a way to access this model for reproducing the results or a way to reproduce the model (\eg, with an open-source dataset or instructions for how to construct the dataset).
            \item We recognize that reproducibility may be tricky in some cases, in which case authors are welcome to describe the particular way they provide for reproducibility. In the case of closed-source models, it may be that access to the model is limited in some way (\eg, to registered users), but it should be possible for other researchers to have some path to reproducing or verifying the results.
        \end{enumerate}
    \end{itemize}

\item {\bf Open access to data and code}
    \item[] Question: Does the paper provide open access to the data and code, with sufficient instructions to faithfully reproduce the main experimental results, as described in supplemental material?
    \item[] Answer: \answerNo{} 
    \item[] Justification: As stated in the response above, we provided detailed instructions on how to replicate our experiment results in the main paper and further in the supplementary material. We will release our code and models upon paper acceptance.
    \item[] Guidelines:
    \begin{itemize}
        \item The answer NA means that paper does not include experiments requiring code.
        \item Please see the NeurIPS code and data submission guidelines (\url{https://nips.cc/public/guides/CodeSubmissionPolicy}) for more details.
        \item While we encourage the release of code and data, we understand that this might not be possible, so “No” is an acceptable answer. Papers cannot be rejected simply for not including code, unless this is central to the contribution (\eg, for a new open-source benchmark).
        \item The instructions should contain the exact command and environment needed to run to reproduce the results. See the NeurIPS code and data submission guidelines (\url{https://nips.cc/public/guides/CodeSubmissionPolicy}) for more details.
        \item The authors should provide instructions on data access and preparation, including how to access the raw data, preprocessed data, intermediate data, and generated data, etc.
        \item The authors should provide scripts to reproduce all experimental results for the new proposed method and baselines. If only a subset of experiments are reproducible, they should state which ones are omitted from the script and why.
        \item At submission time, to preserve anonymity, the authors should release anonymized versions (if applicable).
        \item Providing as much information as possible in supplemental material (appended to the paper) is recommended, but including URLs to data and code is permitted.
    \end{itemize}

\item {\bf Experimental setting/details}
    \item[] Question: Does the paper specify all the training and test details (\eg, data splits, hyperparameters, how they were chosen, type of optimizer, etc.) necessary to understand the results?
    \item[] Answer: \answerYes{} 
    \item[] Justification: We provided detailed instructions on replicating the training and evaluation procedures in Section~\ref{experiment}. We did not perform delicate tuning for the hyperparameters.

\item {\bf Experiment statistical significance}
    \item[] Question: Does the paper report error bars suitably and correctly defined or other appropriate information about the statistical significance of the experiments?
    \item[] Answer: \answerNo{} 
    \item[] Justification: Error bars or statistical significance metrics are not reported, as our experimental protocol follows standard practice in the cross-view geo-localization field, where single-run, fixed split results are widely adopted for benchmarking. For direct comparability, we report results under the same evaluation protocol as prior work. We note that nearly all published baselines on these benchmarks also omit error bars.
    \item[] Guidelines:
    \begin{itemize}
        \item The answer NA means that the paper does not include experiments.
        \item The authors should answer "Yes" if the results are accompanied by error bars, confidence intervals, or statistical significance tests, at least for the experiments that support the main claims of the paper.
        \item The factors of variability that the error bars are capturing should be clearly stated (for example, train/test split, initialization, random drawing of some parameter, or overall run with given experimental conditions).
        \item The method for calculating the error bars should be explained (closed form formula, call to a library function, bootstrap, etc.)
        \item The assumptions made should be given (\eg, Normally distributed errors).
        \item It should be clear whether the error bar is the standard deviation or the standard error of the mean.
        \item It is OK to report 1-sigma error bars, but one should state it. The authors should preferably report a 2-sigma error bar than state that they have a 96\% CI, if the hypothesis of Normality of errors is not verified.
        \item For asymmetric distributions, the authors should be careful not to show in tables or figures symmetric error bars that would yield results that are out of range (\eg negative error rates).
        \item If error bars are reported in tables or plots, The authors should explain in the text how they were calculated and reference the corresponding figures or tables in the text.
    \end{itemize}

\item {\bf Experiments compute resources}
    \item[] Question: For each experiment, does the paper provide sufficient information on the computer resources (type of compute workers, memory, time of execution) needed to reproduce the experiments?
    \item[] Answer: \answerYes{} 
    \item[] Justification: We described our compute setup as well as the  training time and inference runtime in Section~\ref{experiment}.
    \item[] Guidelines:
    \begin{itemize}
        \item The answer NA means that the paper does not include experiments.
        \item The paper should indicate the type of compute workers CPU or GPU, internal cluster, or cloud provider, including relevant memory and storage.
        \item The paper should provide the amount of compute required for each of the individual experimental runs as well as estimate the total compute. 
        \item The paper should disclose whether the full research project required more compute than the experiments reported in the paper (\eg, preliminary or failed experiments that didn't make it into the paper). 
    \end{itemize}
    
\item {\bf Code of ethics}
    \item[] Question: Does the research conducted in the paper conform, in every respect, with the NeurIPS Code of Ethics \url{https://neurips.cc/public/EthicsGuidelines}?
    \item[] Answer:\answerYes{} 
    \item[] Justification: : Authors carefully read the NeurIPS Code of Ethics and preserved anonymity.
    \item[] Guidelines:
    \begin{itemize}
        \item The answer NA means that the authors have not reviewed the NeurIPS Code of Ethics.
        \item If the authors answer No, they should explain the special circumstances that require a deviation from the Code of Ethics.
        \item The authors should make sure to preserve anonymity (\eg, if there is a special consideration due to laws or regulations in their jurisdiction).
    \end{itemize}

\item {\bf Broader impacts}
    \item[] Question: Does the paper discuss both potential positive societal impacts and negative societal impacts of the work performed?
    \item[] Answer: \answerNo{} 
    \item[] Justification: Our paper primarily discusses the positive application prospects of the proposed method, such as urban management and disaster response, in the introduction and application sections. However, we do not specifically address potential negative societal impacts. While the method offers promising practical value, we acknowledge that issues such as privacy protection and ethical risks should be considered during real-world deployment. Future work may further analyze these aspects.
    \item[] Guidelines:
    \begin{itemize}
        \item The answer NA means that there is no societal impact of the work performed.
        \item If the authors answer NA or No, they should explain why their work has no societal impact or why the paper does not address societal impact.
        \item Examples of negative societal impacts include potential malicious or unintended uses (\eg, disinformation, generating fake profiles, surveillance), fairness considerations (\eg, deployment of technologies that could make decisions that unfairly impact specific groups), privacy considerations, and security considerations.
        \item The conference expects that many papers will be foundational research and not tied to particular applications, let alone deployments. However, if there is a direct path to any negative applications, the authors should point it out. For example, it is legitimate to point out that an improvement in the quality of generative models could be used to generate deepfakes for disinformation. On the other hand, it is not needed to point out that a generic algorithm for optimizing neural networks could enable people to train models that generate Deepfakes faster.
        \item The authors should consider possible harms that could arise when the technology is being used as intended and functioning correctly, harms that could arise when the technology is being used as intended but gives incorrect results, and harms following from (intentional or unintentional) misuse of the technology.
        \item If there are negative societal impacts, the authors could also discuss possible mitigation strategies (\eg, gated release of models, providing defenses in addition to attacks, mechanisms for monitoring misuse, mechanisms to monitor how a system learns from feedback over time, improving the efficiency and accessibility of ML).
    \end{itemize}
    
\item {\bf Safeguards}
    \item[] Question: Does the paper describe safeguards that have been put in place for responsible release of data or models that have a high risk for misuse (\eg, pretrained language models, image generators, or scraped datasets)?
    \item[] Answer: \answerNA{} 
    \item[] Justification: Our study does not involve the release of any data or models with a high risk of misuse. All data used in this work are either standard public benchmarks or synthetically generated for research purposes, with no content that requires special safeguards. Additionally, all real-world validation images are collected from publicly available and safe content on Youtube, ensuring no sensitive or inappropriate material is included.

\item {\bf Licenses for existing assets}
    \item[] Question: Are the creators or original owners of assets (\eg, code, data, models), used in the paper, properly credited and are the license and terms of use explicitly mentioned and properly respected?
    \item[] Answer: \answerYes{} 
    \item[] Justification: All assets used in this work, including code, datasets, and models, are properly credited in the text, with full compliance to their respective licenses and terms of use.
    \item[] Guidelines:
    \begin{itemize}
        \item The answer NA means that the paper does not use existing assets.
        \item The authors should cite the original paper that produced the code package or dataset.
        \item The authors should state which version of the asset is used and, if possible, include a URL.
        \item The name of the license (\eg, CC-BY 4.0) should be included for each asset.
        \item For scraped data from a particular source (\eg, website), the copyright and terms of service of that source should be provided.
        \item If assets are released, the license, copyright information, and terms of use in the package should be provided. For popular datasets, \url{paperswithcode.com/datasets} has curated licenses for some datasets. Their licensing guide can help determine the license of a dataset.
        \item For existing datasets that are re-packaged, both the original license and the license of the derived asset (if it has changed) should be provided.
        \item If this information is not available online, the authors are encouraged to reach out to the asset's creators.
    \end{itemize}

\item {\bf New assets}
    \item[] Question: Are new assets introduced in the paper well documented and is the documentation provided alongside the assets?
    \item[] Answer: \answerNA{} 
    \item[] Justification: The paper does not release new assets.

\item {\bf Crowdsourcing and research with human subjects}
    \item[] Question: For crowdsourcing experiments and research with human subjects, does the paper include the full text of instructions given to participants and screenshots, if applicable, as well as details about compensation (if any)? 
    \item[] Answer: \answerNA{} 
    \item[] Justification: The paper does not involve crowdsourcing nor research with human subjects.

\item {\bf Institutional review board (IRB) approvals or equivalent for research with human subjects}
    \item[] Question: Does the paper describe potential risks incurred by study participants, whether such risks were disclosed to the subjects, and whether Institutional Review Board (IRB) approvals (or an equivalent approval/review based on the requirements of your country or institution) were obtained?
    \item[] Answer: \answerNA{} 
    \item[] Justification: The paper does not involve crowdsourcing nor research with human subjects.

\item {\bf Declaration of LLM usage}
    \item[] Question: Does the paper describe the usage of LLMs if it is an important, original, or non-standard component of the core methods in this research? Note that if the LLM is used only for writing, editing, or formatting purposes and does not impact the core methodology, scientific rigorousness, or originality of the research, declaration is not required.
    \item[] Answer: \answerYes{} 
    \item[] Justification: We explicitly describe the use of a large language model (Qwen2.5-VL-32B) as a core component of our methodology. The LLM is leveraged to generate multi-step, chain-of-thought weather and spatial descriptions, which serve as supervisory signals for vision-language alignment in our framework. Detailed descriptions of the LLM's role, prompt design, and integration into the description pipeline are provided in Section~\ref{Method} of the paper.

\end{enumerate}

\end{document}